\documentclass[10pt,twocolumn,letterpaper]{article}

\usepackage[pagenumbers]{cvpr} 
\usepackage{booktabs}
\usepackage{epsfig}
\usepackage{graphicx}
\usepackage{amsmath}
\usepackage{amssymb}
\usepackage{caption}
\usepackage{MnSymbol}
\usepackage{multirow}
\usepackage{pifont}
\usepackage{multicol}
\usepackage{balance}

\usepackage{tabulary}
\usepackage{color, colortbl}
\definecolor{Gray}{gray}{0.95}
\usepackage[dvipsnames]{xcolor}
\usepackage[accsupp]{axessibility}

\usepackage[rightcaption]{sidecap}

\newcommand{\our}{CLAP}

\usepackage{algorithm}
\usepackage{algpseudocode}
\usepackage[dvipsnames]{xcolor}

\definecolor{cvprblue}{rgb}{0.21,0.49,0.74}
\usepackage[pagebackref,breaklinks,colorlinks,citecolor=cvprblue]{hyperref}

\def\paperID{15429} 
\def\confName{CVPR}
\def\confYear{2024}

\title{Collaborative Learning of Anomalies with Privacy (CLAP) for Unsupervised Video Anomaly Detection: A New Baseline}

\author{Anas Al-lahham  \quad Muhammad Zaigham Zaheer \quad Nurbek Tastan \quad Karthik Nandakumar \\ 
Mohamed bin Zayed University of Artificial Intelligence (MBZUAI)\\ 
Abu Dhabi, UAE\\ 
{\tt\small \{anas.al-lahham, zaigham.zaheer, nurbek.tastan, karthik.nandakumar\}@mbzuai.ac.ae}} 

\begin{document}
\maketitle
\begin{abstract}

Unsupervised (US) video anomaly detection (VAD) in surveillance applications is gaining more popularity recently due to its practical real-world applications. As surveillance videos are privacy sensitive and the availability of large-scale video data may enable better US-VAD systems, collaborative learning can be highly rewarding in this setting. However, due to the extremely challenging nature of the US-VAD task, where learning is carried out without any annotations, privacy-preserving collaborative learning of US-VAD systems has not been studied yet. In this paper, we propose a new baseline for anomaly detection capable of localizing anomalous events in complex surveillance videos in a fully unsupervised fashion without any labels on a privacy-preserving participant-based distributed training configuration. Additionally, we propose three new evaluation protocols to benchmark anomaly detection approaches on various scenarios of collaborations and data availability. Based on these protocols, we modify existing VAD datasets to extensively evaluate our approach as well as existing US SOTA methods on two large-scale datasets including UCF-Crime and XD-Violence. All proposed evaluation protocols, dataset splits, and codes are available here: \href{https://github.com/AnasEmad11/CLAP}{https://github.com/AnasEmad11/CLAP}.


\end{abstract}    
\vspace{-15pt}
\section{Introduction}
\label{sec:intro}

Recent years have seen a surge in federated learning based methods, where the goal is to enable collaborative training of machine learning models without transferring any training data to a central server. This direction of research in machine learning is of notable importance as it enables learning with multiple participants that can contribute data without compromising privacy.
Several researchers have studied federated learning for different applications such as medical diagnosis \cite{ng2021federated, almalik2023fesvibs, darzidehkalani2022federated, adnan2022federated}, network security \cite{mothukuri2021survey, liu2020secure, tang2022federated, ghimire2022recent}, and large-scale classification models \cite{bonawitz2019towards, lai2022fedscale, zhang2023towards}. 

\begin{figure}
    \centering
    \includegraphics[width=1\columnwidth]{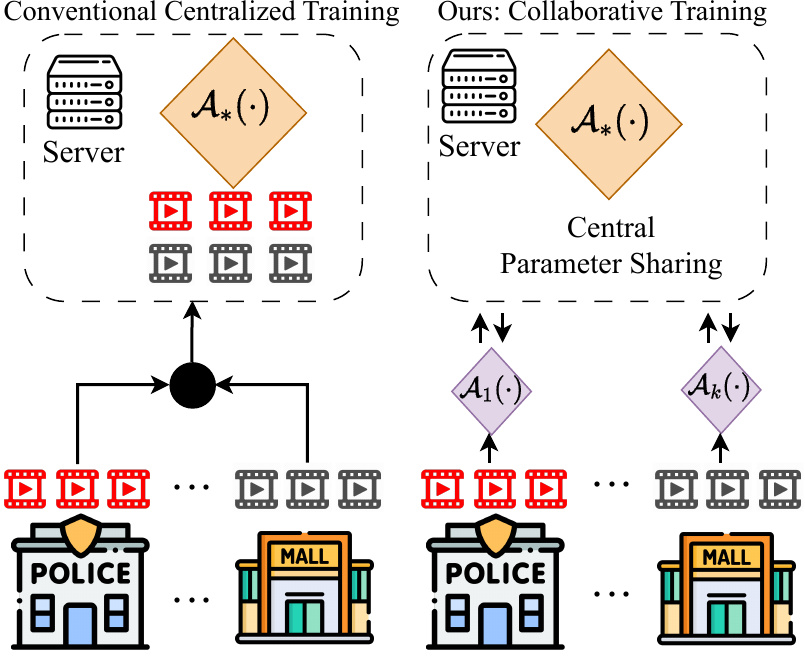}
    \caption{a) Conventional central training requires all training data to be on the server to carry out the training. This setting cannot ensure privacy, thus hindering collaborations between different entities holding large-scale surveillance data. b) Our proposed unsupervised video anomaly detection technique does not require the transfer of training data between the server and participants, thus ensuring complete privacy.}
    \label{fig:intro}
    \vspace{-19pt}
\end{figure}

Anomaly detection in surveillance videos, being one of the large-scale applications of computer vision, may greatly benefit from autonomous collaborative training algorithms. VAD in surveillance videos is privacy sensitive and may involve data belonging to several organizations/establishments. This may result in hectic bureaucratic processes to obtain data from each establishment for centralized training. For example, the police department of a city may not be willing to share street surveillance videos due to privacy concerns of the general public, or a daycare facility may have to obtain the consent of all parents to be allowed to share its CCTV footage.
Such restrictions may hinder the possibility of obtaining large-scale data to train efficient anomaly detectors making a central training requiring all training data the least preferred option in the real-world scenarios.
Unfortunately, to the best of our knowledge, there are hardly any notable attempts to leverage federated learning for video anomaly detection which may be due to the challenging nature of the anomaly detection task itself. Anomalies are often unknown and it is not feasible to collect all possible anomaly examples for a model to learn from. Furthermore, anomalies are rare in nature, and annotating large amounts of data is laborious.

In this work, we explore video anomaly detection (VAD) on two fronts: 1) Unsupervised - videos are used without any labels. 2) Distributed participant  based learning - the server does not get any raw data from participants. Unsupervised VAD is a relatively recent development in the field of anomaly detection in which no supervision is provided during training \cite{zaheer2022generative,anas2023c2fpl}. This class of VAD training is different from the existing one-class classification (OCC) and weakly supervised (WS) learning. In OCC, only normal videos are provided for training 
whereas, in WS both normal and anomalous videos are provided with binary video-level labels. 
Unsupervised VAD is somewhat closer to WS-VAD, as it also utilizes large sets of videos containing normal and anomalous events. However, instead of relying on video-level labels, the network is designed in such a way that it utilizes several cues, such as the abundance of normal events/scarcity of anomalies, etc., present in the surveillance videos to drive the overall training \cite{zaheer2022generative}. Unsupervised VAD itself is a challenging task and the complexity increases multifolds when we consider distributed participants setting for training. However, this is also more rewarding due to zero annotation labor and a more practical real-world application enabling collaboration between different large-scale data sources.

To this end, we propose \our{}, an approach for \underline{C}ollaborative \underline{L}earning of Anomalies with \underline{P}rivacy that takes unlabelled videos at multiple nodes (participants) as input and collaboratively learns to predict frame-level anomaly score predictions as output (Figure \ref{fig:intro}). 
At an abstract level, our approach can be divided into three distinct steps: Common knowledge based data segregation for local training, knowledge accumulation at server, and local feedback.
As we approach the task as fully unsupervised, i.e., without any labels, videos at each participant's end are segregated to separate normal and anomalous candidates. To this end, we propose to utilize Von Neumann entropy as a metric \cite{petz2001entropy}, and apply Gaussian Mixture Model (GMM) to create clusters of normal and anomalous videos. 
After certain local epochs, the weights of all local trainings are accumulated at the server. Based on this, a feedback loop is formed to refine the initial labels obtained by each participant and share the commonly learned knowledge between each participant. 

As video anomaly detection in distributed participant settings is not well-studied, we explore multiple training and evaluation scenarios mimicking real-world collaborations. These scenarios, listed in the order of complexity, include all participants having access to similar type of training data, different participants having access to different types of anomalies, and different participants having totally different types and numbers of videos. 
Overall, the contributions of our manuscript are:
1) We propose, \our{}, a new baseline for anomaly detection capable of localizing anomalous events in complex surveillance scenarios in a fully unsupervised fashion without any labels on a privacy-preserving participant-based training configuration. To the best of our knowledge, \our{} is the first rigorous attempt to tackle video anomaly detection in the federated learning setting.
2) We propose three new evaluation scenarios to extensively evaluate \our{} on various scenarios of collaborations and data availability. 3) To carry out these evaluations, we modify the existing VAD datasets to create new splits.

\section{Related Work}
\label{sec:related_work}

\subsection{Federated Learning}
Federated learning has been studied for various computer vision applications including healthcare \cite{mothukuri2021survey, liu2020secure, tang2022federated}, surveillance \cite{brik2020federated,sada2019distributed,liu2020fedvision,pang2023federated, doshi2023privacy}, and autonomous driving \cite{fantauzzo2022feddrive,du2020federated,zhang2021end,nguyen2022deep,li2021privacy}. Video anomaly detection has not been well-studied in federated learning settings. The closely related anomaly detection in federated learning setting mostly includes network security related methods in which different network attacks are identified from the normal traffic of packets \cite{mothukuri2021survey, liu2020secure, tang2022federated, ghimire2022recent}. However, given that these are mostly supervised tasks in the form of one-class classification, the problem of distributed learning transforms into weight-sharing optimization.
Recently, Doshi \etal \cite{doshi2023privacy} have proposed a weakly supervised federated learning (FL) video anomaly detection (VAD) method. The idea is to explore the FL setting by randomly dividing the data between multiple clients. In essence, this work is related to our approach as we also study FL for VAD. However, we primarily explore the unsupervised setting of VAD. Moreover, without any training labels for VAD, we additionally propose common knowledge-based data segregation for local training and local feedback for improved pseudo-labeling to carry out the collaborative training. Furthermore, we propose realistic scenarios to evaluate VAD methods in FL setting.

\subsection{Unsupervised Anomaly Detection}
Introduced by Zaheer \etal \cite{zaheer2022generative}, fully unsupervised anomaly detection is a relatively new idea and the methods that do not require any training labels are still quite sparse in the literature. 
This problem is extremely challenging due to the rarity of anomalies and the complete lack of supervision labels. 
Zaheer \etal \cite{zaheer2022generative}, by relying on the abundance of normal data, proposed to first train a generative model to learn the overall normal trends in the dataset. A classifier model is then trained based on the pseudo labels obtained from the generator. Both models are then trained in a cooperative manner to converge as an anomaly detector.
This line of research has been extended by Anas \etal \cite{anas2023c2fpl}
. They have proposed to utilize hierarchical clustering to obtain fine-grained pseudo-labels. The training of an anomaly detector is then carried out using these pseudo labels. 
Our approach also begins with pseudo-label generation, followed by training and then a feedback loop to improve the pseudo-labels. However, we attempt to address the problem in collaborative learning setting where privacy-preserving training is carried out by multiple participants. In addition, we propose a common knowledge aware data segregation in which all participants share common clusters knowledge to obtain pseudo-labels for training.

We also acknowledge one-class classification (OCC) \cite{zaheer2020old,sabokrou2018ALOCC,gong2019memorizing,ionescu2019objectcentric} and weakly-supervised (WS) \cite{sultani2018real,zaheer2020claws,zaheer2020self,tian2021featuremagnitude,wu2020not,zhang2023exploiting} approaches for video anomaly detection. To enrich the quality of analysis and to provide several aspects of our research work, we carry out some additional experiments on weakly-supervised settings by incorporating weak labels in our training in Section \ref{sec:experiments}. However, the prime focus of our research work is unsupervised video anomaly detection, and thus, OCC and WS anomaly detection are not in the scope of our research work.

\section{Methodology}
\label{sec:methodology}
\noindent \textbf{Problem Definition}: Given a dataset of training videos without any labels, the goal of US-VAD is to learn an anomaly detector $\mathcal{A}_{\theta}(\cdot)$ that classifies each frame in a given test video $V_*$ as either \textit{normal} ($0$) or \textit{anomalous} ($1$). Suppose that there are $K$ participants $\mathcal{P}_{1}, \mathcal{P}_{2}, ..., \mathcal{P}_{K}$ for collaborative training and each participant $\mathcal{P}_{k}$ has its own local training dataset $\mathcal{D}_{k} = \{V_{1,k}, V_{2,k}, \cdots, V_{N_k,k}\}$ containing $N_k$ videos used to train its own local anomaly detector model  $\mathcal{A}_{\theta_k}(\cdot)$, $k \in [1,K]$. It is assumed that all participants share a common test set and the performance of the model $\mathcal{A}_{\theta_k}(\cdot)$ on this test set is denoted by $\mathcal{O}_k$. The goal of each participant is to collaborate with other participants in order to obtain a global model $\theta_{*}$ that has better performance $\mathcal{O}_{*}$ on the test set compared to all $\mathcal{O}_k$, without compromising the privacy of participant's local data $\mathcal{D}_k$.
  
\begin{figure}[t]
    \centering
    \includegraphics[width=1\linewidth]{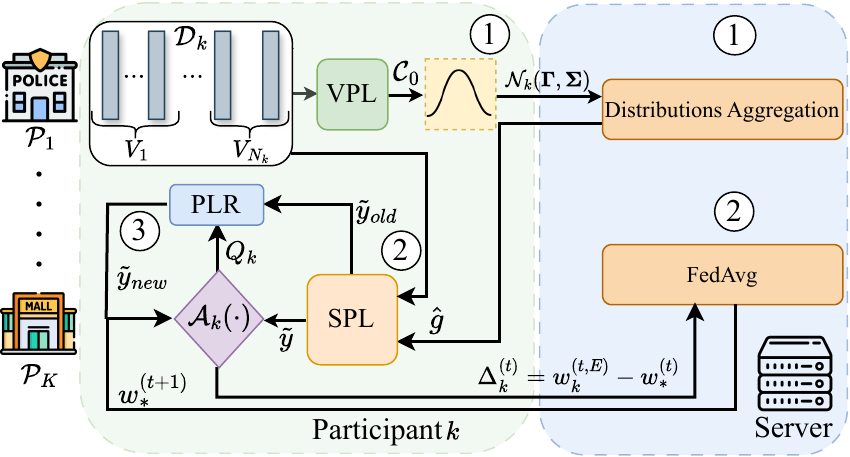}
    \caption{Architecture of \our, an unsupervised video anomaly detection model trained by multiple collaborating participants.}
    \label{fig: pseudo-labeling}
\vspace{-15pt}
\end{figure}

\noindent \textbf{Preprocessing}: For simplicity of notation, we drop the participant index $k$ unless required. Let the training dataset of $N$ videos at a generic participant be denoted as $\mathcal{D} = \{V_{1}, V_{2}, \cdots, V_{N}\}$. We split each video $V_i$ into a sequence of $m_{i}$ non-overlapping segments $S_{ij}$, where each segment is in turn composed of $r$ frames. Note that $i \in [1,N]$ refers to the video index and $j \in [1,m_{i}]$ is the segment index within a video. Unlike many state-of-the-art AD methods \cite{sultani2018real,tian2021weakly,wu2022self, Purwanto2021ICCV} that compress each video into a fixed number of segments (i.e., $m_i =m, \forall i \in [1,N]$) along the temporal axis, we avoid any compression and make use of all available non-overlapping segments \footnote{Note that for simplicity, we use the notation $val_{ij}$ to represent a value for segment $j$ in video $V_i$, $val_{i}$ represents the set of values of all segments in video $V_i$, and $val$ simply represents the collection of all segment-wise values of all the videos in the dataset $\mathcal{D}$.}. For each segment $S_{ij}$, a feature vector $\mathbf{f}_{ij} \in \mathbb{R}^{d}$ is obtained using a pre-trained feature extractor $\mathcal{F}(\cdot)$. 


\noindent \textbf{High-level Overview}:
Our proposed \our{} model for collaborative training consists of three main stages. The aim is to generate both video-level and segment-level pseudo-labels to enable the training of the anomaly detector model $\mathcal{A}_{\theta}(\cdot)$. In the first Common Knowledge-based Data Segregation (CKDS) stage, the generation of segment-level pseudo labels is done in a collaborative manner. We generate a video-level pseudo-label $\hat{y}_i \in \{0,1\}$, $i \in [1,N]$ for each video in the training set using hierarchical divisive clustering. Subsequently, segment-level pseudo-labels $\tilde{y}_{ij} \in \{0,1\}$, $i \in [1,N]$, $j \in [1,m_{i}]$ are generated for all the segments in the training set through collaborative statistical hypothesis testing. Finally, we train a local anomaly detector $\mathcal{A}_{\theta}(\cdot): \mathbb{R}^{d} \rightarrow [0,1]$ that assigns an anomaly score between $0$ and $1$ (higher values indicate higher confidence of being an anomaly) to the given video segment based on its feature representation $\mathbf{f}_{ij}$.

During training, we utilize both server knowledge accumulation (SKA) and local feedback stages to improve the performance of the local model. In the second (SKA) stage, we use the well-known Federated Averaging (FedAvg) algorithm \cite{mcmahan2017communication} (we also analyze other FL aggregation methods in Section \ref{otherfl} of supplementary material) for aggregating local anomaly detection models. Finally, upon completion of a pre-determined number of collaboration rounds to update the weights given the initial pseudo-labels, our algorithm initiates the local feedback or pseudo-label refinement (PLR) stage. During this stage, we use confidence scores predicted by the network to refine the generated pseudo-labels from the first stage.


\subsection{Knowledge-based Data Segregation (CKDS)}
\label{sec:CKDS}
At the participant level, since the training does not assume any labels, we first generate pseudo-labels for the videos in the training set by clustering them into two groups: normal and anomalous (see Alg. \ref{alg:cpl}).

\noindent \textbf{Video-Level Pseudo-Labels}:
Previous works in WS-VAD have shown that normal video segments have lower temporal feature magnitude compared to anomalous segments \cite{tian2021featuremagnitude}. In addition, we observe the variance of the difference in the feature magnitude between consecutive segments in a given anomalous video is higher than in a normal video. 
Furthermore, we consider von-Neumann entropy $H$ of the covariance matrix computed based on the features as an indicator of the presence of anomalies, i.e., the entropy of segments is generally expected to be lower for normal videos. Based on these cues, we represent each video $V_i$ using a statistical summary $\mathbf{x}_i = [\sigma_i, H_i]$ of its features as follows:

\begin{equation}
    \mu_i = \frac{1}{(m_i-1)} \sum_{j=1}^{(m_i-1)} \left(||\mathbf{f}_{ij}||_2 - ||\mathbf{f}_{i(j+1)}||_2\right), 
    \label{eq:mu}
\end{equation}

\begin{equation}
    \sigma_i = \sqrt{\frac{1}{(m_i-2)} \sum_{j=1}^{(m_i-1)}((||\mathbf{f}_{ij}||_2 - ||\mathbf{f}_{i(j+1)}||_2) - \mu_i)^2}, 
    \label{eq:sigma}
\end{equation}

\begin{equation}
    \text{Cov}[\mathbf{f}_{i,1}, ..., \mathbf{f}_{i,m_i}] = \mathbf{U}_i \mathbf{\Sigma}_i \mathbf{V}_i^T 
    \label{eq:cov}
\end{equation}

\begin{equation}
    H_i = -\text{tr} [\mathbf{\Sigma}_i \log \mathbf{\Sigma}_i] 
    \label{eq:entropy}
\end{equation}

\noindent  where $||\cdot||_2$ represents the $\ell_2$ norm of a vector.  
Thus, each video $V_i$ is represented using a 2D vector $\mathbf{x}_i$, corresponding to the variance $\sigma_i$ and entropy $H_i$ of the video segments. Videos in the training set are then divided into two clusters ($\mathcal{C}_{s_{1}}$ and $\mathcal{C}_{s_{2}}$), with 
$|\mathcal{C}_{s_{1}}|$ and $|\mathcal{C}_{s_{2}}|$ samples, respectively, based on the above representation $\mathbf{x}_i$. 

Intuitively, building on the assumption that normal samples have a lower entropy, we compute the average entropy of samples in each cluster and assign the cluster with the larger average entropy as anomalous ($0$), while the other cluster is labeled as normal ($1$). At the end of this stage, all the videos in the training set are assigned a pseudo-label based on their corresponding cluster label, i.e., $\hat{y}_i = s$, if $\mathbf{x}_i \in \mathcal{C}_{s}$, where $s \in \{0,1\}$.


\begin{algorithm}[t]
\caption{Video-level Pseudo-Label Generation (VPL)}
\label{alg:cpl}
\begin{algorithmic}[1] 
\Statex \textbf{Input:} dataset $\mathcal{D} = \{V_1, \cdots , V_N\}$, feature extractor $\mathcal{F}(\cdot)$

\For{$i = 1$ to $N$} 
    \State Partition $V_i$ into $m_i$ segments $[S_{i1},\cdots,S_{im_i}]$
    \State Extract segment features $[\mathbf{f}_{i1},\cdots, \mathbf{f}_{im_i}]$ using $\mathcal{F}(\cdot)$
    \State Compute $\mathbf{x}_i = [\sigma_i, H_i]$ using Eqs. \ref{eq:sigma} \& \ref{eq:entropy} 
\EndFor 
\State $\mathcal{C}_{s_{1}}$, $\mathcal{C}_{s_{2}}$ $\gets$ Clustering(GMM), where $a = |\mathcal{C}_{s_{1}}|, b = |\mathcal{C}_{s_{2}}| $
\If  {$ \frac{1}{a}\sum_{i=1}^{a} H_{i} > \frac{1}{b}\sum_{i=1}^{b} H_{i} $}
    \State $\mathcal{C}_0$  = $\mathcal{C}_{s_{2}}$ , $\mathcal{C}_1$  = $\mathcal{C}_{s_{1}}$
\Else
    \State $\mathcal{C}_0$  = $\mathcal{C}_{s_{1}}$ , $\mathcal{C}_1$  = $\mathcal{C}_{s_{2}}$
\EndIf

\State $\forall i \in [1,N], \hat{y}_i \gets 0$ \textbf{if} {$\mathbf{x}_i \in \mathcal{C}_0$}, \textbf{else} $\hat{y}_i \gets 1$
\Statex \textbf{return} $\hat{y} = \{\hat{y}_1,\cdots,\hat{y}_N\}$ 
\end{algorithmic}
\end{algorithm}

\noindent \textbf{Segment-Level Pseudo-Labels}: 
All the segments from videos that are ``pseudo-labeled'' as normal ($\hat{y}_i=0$) by the previous stage can be considered normal. However, most of the segments in an anomalous video are also normal due to the smaller temporal extent of anomalies.  To tackle this, we treat the detection of anomalous segments as a statistical hypothesis-testing problem. Specifically, the null hypothesis is that a given video segment is normal. By modeling the distribution of features under the null hypothesis as a Gaussian distribution, we identify the anomalous segments by estimating their p-value and rejecting the null hypothesis if the p-value is less than the significance level $\alpha$.

\noindent To model the distribution of features (\textbf{at the participant level}) under the null hypothesis, we consider only the segments from videos that are pseudo-labeled as normal by the VPL stage (see Algo. \ref{alg:cpl}). Let $\mathbf{z}_{ij} \in \mathbb{R}^{\tilde{d}}$ be a low-dimensional representation of a segment $S_{ij}$. In this work, we simply set $\mathbf{z}_{ij} = ||\mathbf{f}_{ij}||_2$. We assume that $\mathbf{z}_{ij}$ follows a Gaussian distribution $\mathcal{N}(\mathbf{\Gamma},\mathbf{\Theta})$ under the null hypothesis and estimate the parameters $\mathbf{\Gamma}$ and $\mathbf{\Theta}$ as follows:

\begin{equation}
    \mathbf{\Gamma} = \frac{1}{M_0} \sum_{i=1, \hat{y}_i = 0}^{N}\sum_{j=1}^{m_i}\mathbf{z}_{ij}, 
    \label{eq:gamma}
\end{equation}

\begin{equation}
    \mathbf{\Theta} = \frac{1}{(M_0-1)} \sum_{i=1, \hat{y}_i = 0}^{N}\sum_{j=1}^{m_i} (\mathbf{z}_{ij} - \mathbf{\Gamma})(\mathbf{z}_{ij} - \mathbf{\Gamma})^T, 
    \label{eq:Sigma}
\end{equation}

\noindent where $M_0 = \sum_{i=1, \hat{y}_i = 0}^N m_i$. 

Let $(\mathbf{\Gamma}_k,\mathbf{\Theta}_k)$ be the Gaussian parameters at participant $\mathcal{P}_{k}$ based on $M_{0,k}$ normal segments. The participants share these parameters with the server and the server sends back a Gaussian mixture model $\mathcal{G}$. One simple way to construct $\mathcal{G}$ is to treat $(\mathbf{\Gamma}_k,\mathbf{\Theta}_k)$  of each participant as a mixture component and weight these components based on the corresponding number of normal segments. More sophisticated aggregation approaches could also be employed by the server. Subsequently, for all the segments in videos that are pseudo-labeled as anomalous, the $p$-value is computed as:

\begin{equation}
    \label{eq: pdf}
          p_{ij} = \mathcal{G}({\mathbf{z}}_{ij}), ~\forall ~ \hat{y}_i = 1.
\end{equation}


\noindent If $p_{ij} < \alpha$, the segment can be directly assigned a pseudo-label of $1$. However, we identify a potential anomalous region by sliding a window of size $w_i$ across the video and selecting the region that has the lowest average p-values (i.e., $\min_{l} \left\{ \frac{1}{w_i} \sum_{j=(l+1)}^{(l+w_i)} p_{ij}, ~ \forall ~ l \in [0,m_i-w_i] \right\}$). Each segment present in this anomalous region is assigned a pseudo-label of $1$, while all the remaining segments are pseudo-labeled as normal with a value of $0$. Thus, a pseudo-label $\tilde{y}_{ij} \in \{0,1\}$ is assigned to all the segments in the training set. This window-based labeling may be seen as utilizing the temporal consistency property of the surveillance videos commonly utilized in the existing literature \cite{sultani2018real,zaheer2020claws}.


\begin{algorithm}
\caption{Segment-level Pseudo-Label Gen. (SPL)}
\label{alg:fpl}
\begin{algorithmic}[1] 
\Statex \textbf{Input:} Mode, $0 < \beta \leq 1$, $\hat{y}$ (video-level pseudo-labels) and Gaussian mixture model $\mathcal{G}$ for Generate mode, $\tilde{y}$ (current segment-level pseudo-labels) and $Q$ (segment-level confidence scores) for Update mode 


\Statex \textbf{Mode I:} Generate  
\For{$i = 1$ to $N$} 
    \If{$\hat{y}_i = 1$}
        \State Compute $p_{ij}$ using Eq. \ref{eq: pdf}, $\forall j \in [1,m_i]$
        \State $w_i \gets \lceil\beta m_i\rceil$
        \State $l_i  = \arg\min_{l} \left\{ \frac{1}{w_i} \sum_{j=(l+1)}^{(l+w_i)} p_{ij}, ~ \forall l \in [0,m_i-w_i] \right\}$ 
        \State $\tilde{y}_{ij} \gets 1, ~\forall j \in [l_i+1,l_i+w]$ 
    \EndIf
\EndFor
\Statex \textbf{return} $\tilde{y}$ 
\Statex \textbf{Mode II:} Update (PLR)
\For{$i = 1$ to $N$}
    \State  Set $q_{ij}$ based on $Q~ \forall ~ j \in [1, m_i]$
    \State $w_i \gets \lceil\beta m_i\rceil$
    \State $l_i  = \arg\max_{l} \left\{ \frac{1}{w_i} \sum_{j=(l+1)}^{(l+w_i)} q_{ij}, ~ \forall l \in [0,m_i-w_i] \right\}$ 
    \State $\tilde{q}_{ij} \gets 0, ~\forall j \in [1, m_i]$ 
    \State $\tilde{q}_{ij} \gets 1, ~\forall j \in [l_i+1,l_i+w]$ 
    \State $\hat{q}_{ij} \gets 0, ~\forall j \in [1, m_i]$
    \If{$\text{len}(\tilde{y}_{i} \cap \tilde{q}_{i}) > 0$}
        \State $\hat{q}_{ij} \gets 1, ~ \forall j \in [1, m_i], ~\text{if } ~ \tilde{q}_{ij} \in (\tilde{y}_i \cap \tilde{q}_i)$ 
    \Else 
        \State $\hat{q}_{ij} \gets 1, ~ \forall j \in [1, m_i], ~\text{if } ~ \tilde{q}_{ij} \neq 0$ or $\tilde{y}_{i,j} \neq 0$ 
    \EndIf
\EndFor
\State \textbf{return} $\hat{q}$ 

\end{algorithmic}
\end{algorithm}

\begin{algorithm}
\caption{\our}
\label{alg:flad}
\begin{algorithmic}[1]
\Statex \textbf{Require:} Local training dataset $\mathcal{D}_{k}$. Server initializes parameter $\theta_*^{(0)}$.

\For{each participant $k = 1,2,..., K$} 
    \State $\hat{y_k} \gets $ \textbf{Algorithm \ref{alg:cpl}} ($\mathcal{D}_{k}$)
    \State $\forall i \in [1,N_k], j \in [1,m_i]$, $\tilde{y}_{ij} \gets 0$, Compute $\mathbf{z}_{ij}$
    \State Compute $(\mathbf{\Gamma_k}, \mathbf{\Theta_k})$ using Eqs. \ref{eq:gamma} \& \ref{eq:Sigma} 
    \State \textbf{return} $(\mathbf{\Gamma_k}, \mathbf{\Theta_k})$ to server 
    \EndFor
    \State server: $\mathcal{G} \leftarrow$ Mixture of Gaussians ($\{
   (\mathbf{\Gamma_k},\mathbf{\Theta_k})\}_{k=1}^K$)
\For{each participant $k = 1,2,..., K$}
    \State $\tilde{y_k} \gets $ \textbf{Algorithm \ref{alg:fpl}}  (Generate,  $\hat{y_k}$, $\mathcal{G}$)
\EndFor

\For{each round $ t = 0,1,..., T$} 
    \For{each participant $k = 1,2,..., K$} 
    \State $\tilde{\mathcal{D}}_k \gets (\mathcal{D}_k,\tilde{y}_k)$
    \State $\theta_k^{(t, 0)} \gets \theta_*^{(t)}$ 
    \For{local iteration $e = 0,1,..., E$} 
    \State $\theta_k^{(t, e+1)} \gets \theta_k^{(t, e)} - \eta \cdot \nabla \mathcal{L}_{total,k}(\tilde{\mathcal{D}}_k,\theta_k^{(t, e)}) $ 
    \EndFor 
    \State $\Delta_k^{(t)} \gets \theta_k^{(t,E)} - \theta_*^{(t)}$
    \State $Q_k \gets \mathcal{A}_{\theta_k^{(t,E)}}(\mathcal{D}_k)$
     \State $\tilde{y_k} \gets $ \textbf{Algorithm \ref{alg:fpl}}  (Update, $Q_{k}$, $\tilde{y_k}$)
    \State \textbf{return} $\Delta_k^{(t)}$ to server 

    \EndFor 
    
    \State server: $\theta_*^{(t+1)} \gets \theta_*^{(t)} + \frac{\lambda}{K} \sum_{k=1}^{K} \Delta_k^{(t)}$ 

    
\EndFor

\end{algorithmic}
\end{algorithm}


\subsection{Server Knowledge Accumulation (SKA) and Local Feedback}
At the beginning of this stage, each participant would have obtained segment-level pseudo-labels for its own dataset by sequentially applying Algo. \ref{alg:cpl} and Algo. \ref{alg:fpl} as described in Section \ref{sec:CKDS}.
As earlier, if $\mathcal{D}$ is the unlabeled training dataset of a generic participant, we can obtain the segment-level pseudo-labeled training set $\tilde{\mathcal{D}} = \{(\mathbf{f}_{ij},\tilde{y}_{ij})\}$ containing $M$ samples, where $i \in [1,N]$, $j \in [1,m_{i}]$, and $M = \sum_{i=1}^n m_{i}$. This labeled training set $\tilde{\mathcal{D}}$ can be used to train the anomaly detector $\mathcal{A}_{\theta}(\cdot)$ by minimizing the following objective:

\begin{equation}
    \min_{\theta} ~ \mathcal{L}_{total} = \sum_{i=1}^{N}\sum_{j=1}^{m_{i}} \mathcal{L}(\mathcal{A}_{\theta}(\mathbf{f}_{ij}),\tilde{y}_{ij}),
\end{equation}

\noindent where $\mathcal{L}$ is an appropriate loss function and $\theta$ denotes the parameters of the anomaly detector $\tilde{\mathcal{A}}(\cdot)$. Stochastic gradient descent (SGD) is used for the above optimization. Following recent state-of-the-art methods \cite{zaheer2022generative,tian2021weakly, zaheer2020claws}, 
a basic neural network architecture is considered for our anomaly detector (see Supplementary for more details). 

Now, we proceed to describe the collaborative training of the anomaly detector, which is referred to as \textbf{server knowledge accumulation} (SKA). At the beginning of each collaboration round $t$, the server broadcasts the current global model parameters $\theta_{*}^{(t)}$ to all the participants. Using these global parameters as the initialization, each participant will perform $E$ local SGD iterations to get the updated parameters $\theta_k^{(t,E)}$. At the end of the local training, the participant sends the local gradient $\Delta_k^{(t)} \gets \theta_k^{(t,E)} - \theta_*^{(t)}$ back to the server. The server aggregates these gradients and applies the update to the global model as $\theta_*^{(t+1)} \gets \theta_*^{(t)} + \frac{\lambda}{K} \sum_{i=1}^{K} \Delta_k^{(t)}$ as shown in Algo. \ref{alg:flad}, where $\lambda$ is the learning rate.

In addition to SKA, we also incorporate local feedback or \textbf{pseudo-label refinement process} (PLR) as shown in Algo.\ref{alg:fpl}(Update). During this stage, we use confidence scores $Q$ predicted by the local model to refine the segment-level pseudo-labels. The aim is to use the high-confidence segments to update the pseudo-labels $\tilde{y}$ generated from Algo.\ref{alg:fpl}(Generate). First, we determine the maximum confidence region by a sliding window $w_i$ similar to Algo.\ref{alg:fpl}(Generate) and assign those segments as $\tilde{q_{ij}} = 1$. The refinement of the old pseudo-labels is based on two rules. First, if there is an intersection between the maximum confidence region and the generated pseudo-labels $(\tilde{y}_{i} \cap \tilde{q}_{i}) > 0$, we assign all the segments in $\tilde{q}_{ij}$ that are in the intersection set a value of 1. On the other hand, if there is no intersection, we assume that the old pseudo-labels missed an additional anomalous window. Therefore, the whole of the maximum confidence region will be assigned as anomalous.

\subsection{Inference} 
At the end of all communication rounds, the final global model $\mathcal{A}_{\theta_{*}^{(T)}}(\cdot)$ is used for inference. A given test video $V_*$ is partitioned into $m_*$ non-overlapping segments $S_{*j}$, $j \in [1,m_*]$. Feature vectors $\mathbf{f}_{*j}$ are extracted from each segment using $\mathcal{F}(\cdot)$, which are directly passed to the trained detector $\mathcal{A}_{\theta_{*}^{(T)}}(\cdot)$ to obtain segment-level anomaly score predictions. As the final goal is frame-level anomaly scores, all frames within a segment of the test video inherit the predicted anomaly score for that corresponding segment.


%


\section{Experiments and Results}
\label{sec:experiments}

\begin{figure*}[t]
    \centering
    
    \begin{subfigure}[b]{0.29\textwidth}
        \includegraphics[width=\linewidth]{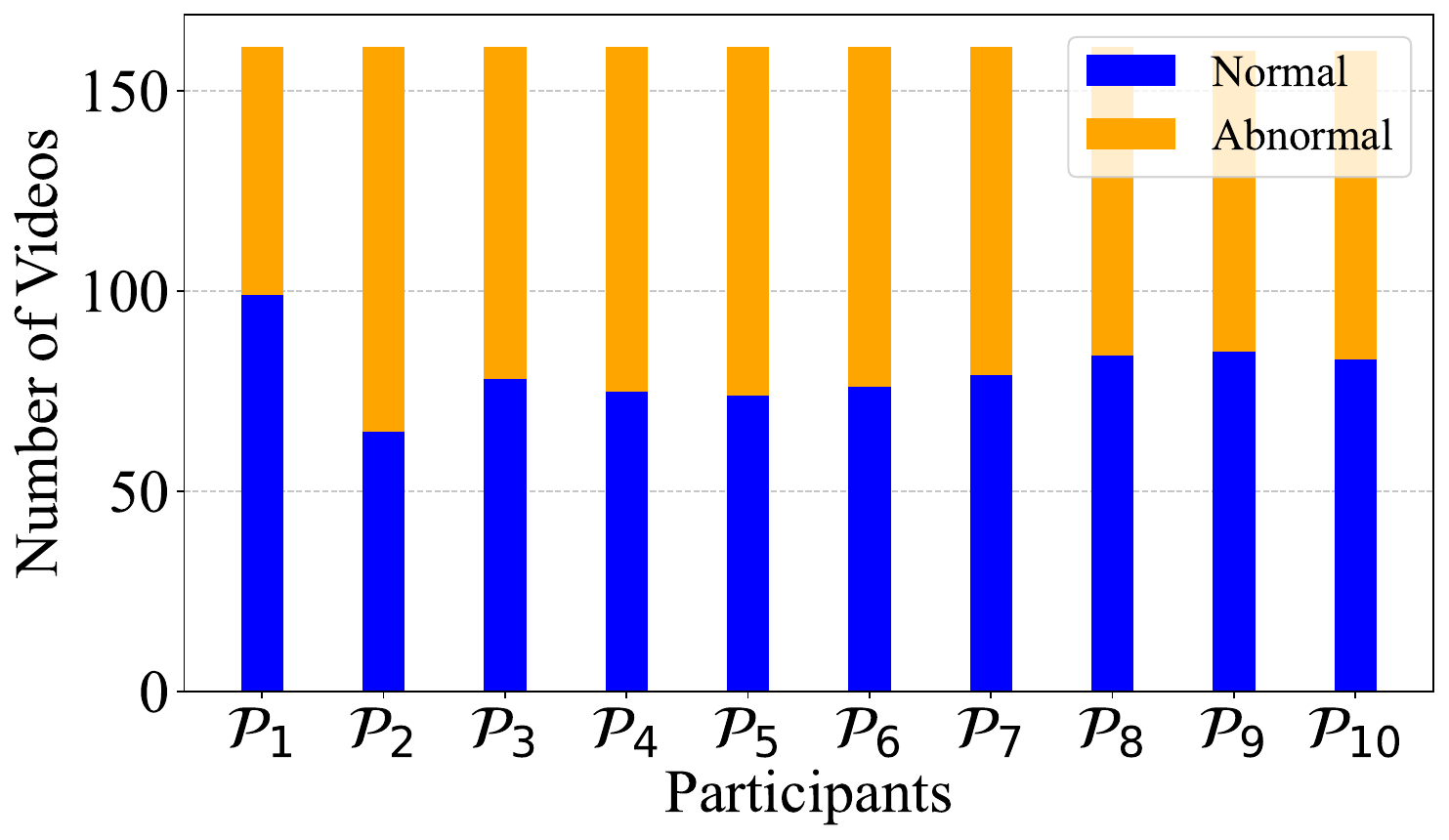}
        \caption{Random  Distribution}
        \label{fig:subfig1}
    \end{subfigure}
    \hfill
    \begin{subfigure}[b]{0.3\textwidth}
        \includegraphics[width=\linewidth]{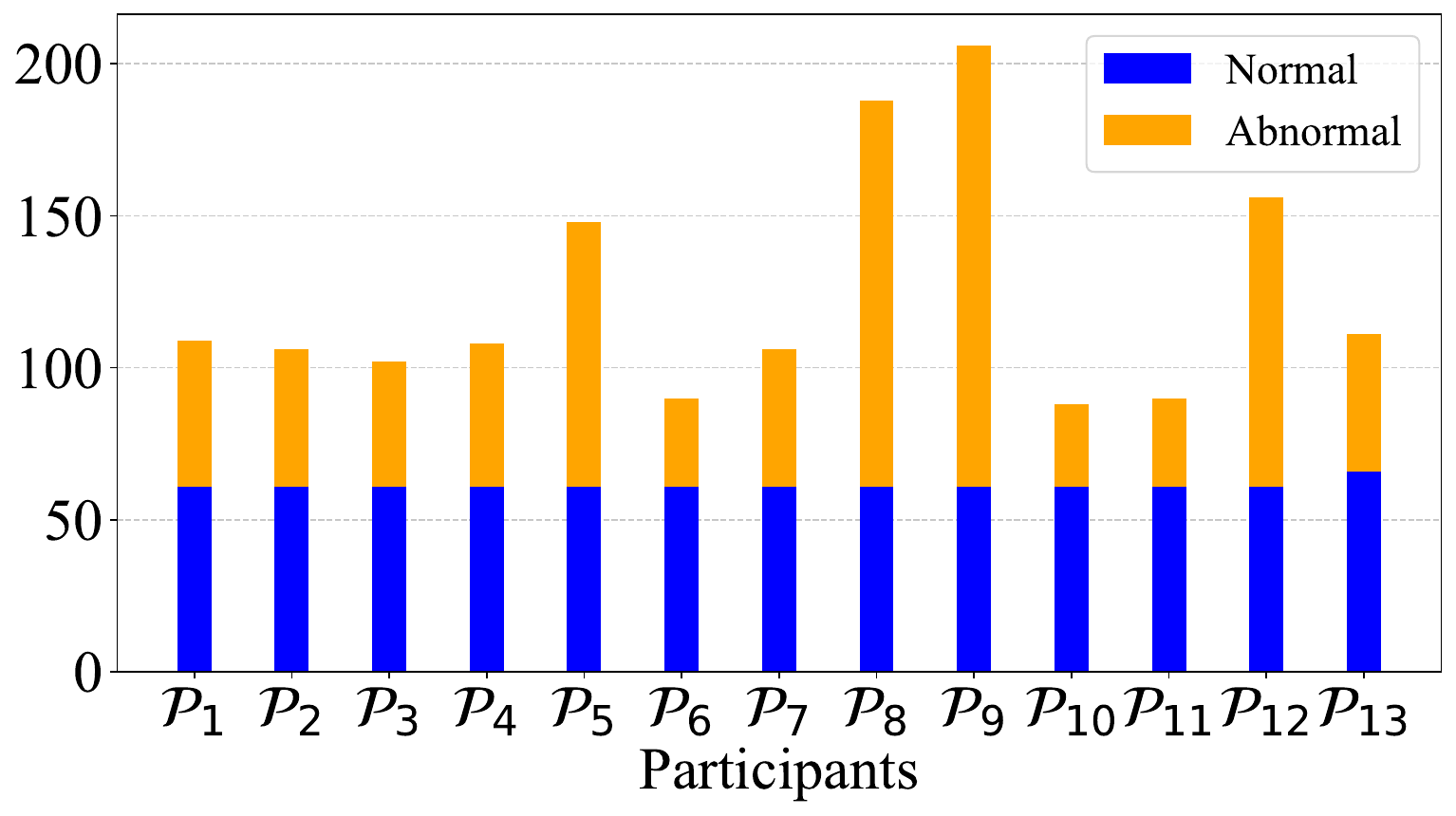}
        \caption{Event Based Distribution}
        \label{fig:subfig2}
    \end{subfigure}
    \hfill
    \begin{subfigure}[b]{0.3\textwidth}
        \includegraphics[width=\linewidth]{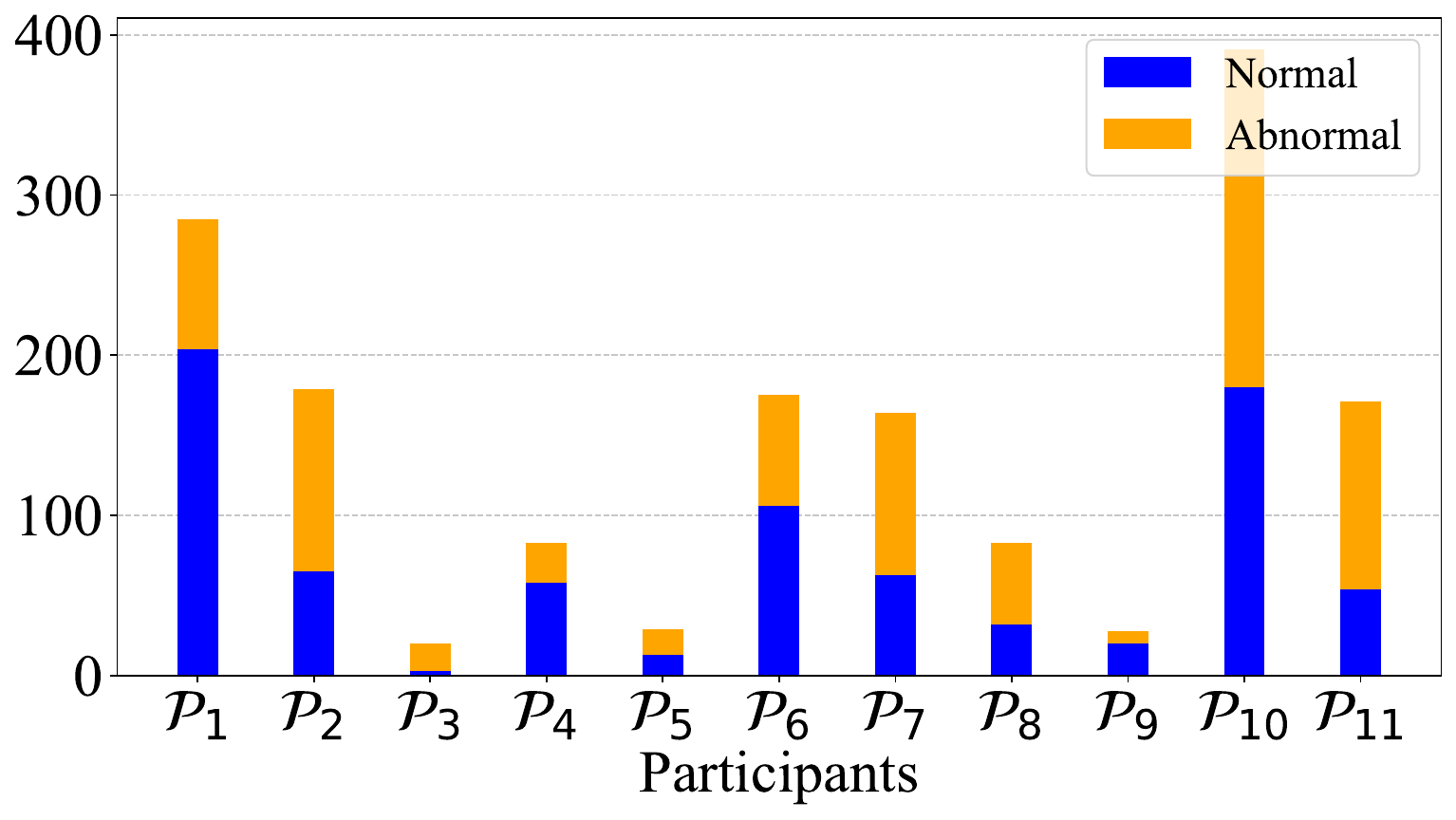}
        \caption{Scene Based  Distribution}
        \label{fig:subfig3}
    \end{subfigure}
    
    \caption{Distribution of UCF-Crime dataset videos based on the three training data organizations proposed in our paper to evaluate collaborative learning approaches for video Anomaly Detection. (a) Random distribution of the videos is the baseline in which each participant has an almost identical number of videos and the classes are balanced. (b) Each participant holds videos containing certain types of anomalous events such as. shooting, robbery, etc. It is a relatively complex setting with the number of videos and class balance varying slightly between participants.  (c) Each participant holds videos belonging to certain scenes such as shops, offices, etc. This is the most challenging setting where severe data and class imbalance are present across participants. }
    \label{fig:training_data_organization}
\end{figure*}


\subsection{Datasets and Implementation}
We evaluate \our{} and conduct SOTA comparisons on two publicly available large-scale datasets, UCF-Crime and XD-Violence. Due to limited space, datasets and implementation details are provided in Supplementary.

    
  


\subsection{Training Settings}

As the aim of \our{} is to train an anomaly detector with collaboration between multiple participants, we provide comparisons of our approach with existing SOTA anomaly detection methods under the following three different training settings:

\noindent\textbf{Centralized Training:} This is the conventional setting of training where privacy is not ensured and the participants have to send all training data to the server for joint training. Anomaly detection performance is measured on the complete test set.\\
\noindent\textbf{Local Training:} This setting assumes that participants are not collaborating and each individual participant trains its anomaly detector locally with its own data. Anomaly detection performance is measured for each participant individually on the complete test set.\\
\noindent\textbf{Collaborative Training:} 
In this setting, all participants collaborate to train a joint anomaly detector. Participants do not need to send their training data to the server to carry out the training. Anomaly detection performance of the jointly trained model is measured on the complete test set.

\renewcommand{\arraystretch}{1.1}
\begin{table}[t]
\scriptsize
\begin{center}
\begin{tabulary}{\linewidth}{CCCC}
\rowcolor{lightgray}
\toprule
               & Method  & UCF-Crime   & XD-Violence \\ \midrule
\multirow{3}{*}{Centralized}     
               & GCL \cite{zaheer2022generative}     &   71.04        &   73.62 \\
               & C2FPL \cite{anas2023c2fpl}   & 80.65        & 80.09   \\
               & \our{}    & \textbf{80.9}    & \textbf{81.71} \\
\midrule
\multirow{3}{*}{Local }           
               & GCL \cite{zaheer2022generative}     &    56.63   &   58.11 \\
               & C2FPL \cite{anas2023c2fpl}   &  61.33       & 60.05   \\
               & \our{}    & \textbf{63.93}           & \textbf{62.37}   \\
 \midrule
\multirow{3}{*}{Collaborative } 
               & GCL \cite{zaheer2022generative}     &  67.12      & 68.19   \\
               & C2FPL \cite{anas2023c2fpl}  &  75.20       & 74.36   \\
                & \our{}    & \textbf{78.02}      &   \textbf{77.65}   \\
\bottomrule
\end{tabulary}
\end{center}
\vspace{-15pt}
\caption{AUC performance comparisons of unsupervised SOTA on UCF-Crime and XD-Violence datasets for five participants.}
\label{tab:us_sota_comp}
\vspace{-18pt}

\end{table}

\subsection{Comparisons with Unsupervised SOTA}

Table \ref{tab:us_sota_comp} summarizes the results of the existing unsupervised anomaly detection approaches on the three training settings: centralized, local, and collaborative. We re-implemented the existing methods \cite{zaheer2022generative,anas2023c2fpl} on local and collaborative training settings for fair and detailed comparisons. 
In essence, centralized training is the upper bound of the collaborative training whereas local training is the lower bound. While better performance compared to the local training setting may indicate the success of collaborative learning, a better model should also demonstrate minimal performance difference from the centralized training.
\our{} demonstrates AUC performances of 80.91\% and 81.71\% on UCF-crime and XD-Violence datasets on the centralized setting (upper bound). In the local setting (lower bound) with five participants, \our{} demonstrate AUC performances of 63.93\% and 62.37\% on the two datasets. In the collaborative learning setting, \our{} demonstrates AUC performances of 78.02\% and 77.65\%. Overall, the results of \our{} are not only better than the existing SOTA unsupervised VAD methods but are also on par with its counterpart centralized training setting.





\subsection{Comparisons with Weakly-supervised SOTA}

An unsupervised method can be converted into a supervised method upon the availability of labels \cite{zaheer2022generative}. We explore this supervision mode under centralized, local, and collaborative training settings and report the results in Table \ref{tab:ws_sota_comp}. Compared to the methods developed specifically for unsupervised video anomaly detection \cite{zaheer2022generative,anas2023c2fpl}, \our{} demonstrates consistent performance improvements when video-level labels are present. \our{} also outperforms PRV \cite{doshi2023privacy}, a weakly-supervised approach designed specifically for collaborative learning. 
Compared to other centralized approaches that do not facilitate unsupervised training, \our{} demonstrates better performance than the compared methods on XD-Violence dataset and comparable performance on UCF-Crime dataset. Nevertheless, the goal of this work is not to surpass performance numbers on certain tasks but to demonstrate the possibility of unsupervised training under a collaborative learning setting to facilitate a novel research direction in the field of video anomaly detection.


\renewcommand{\arraystretch}{1.1}
\begin{table}[t]
\scriptsize
\begin{center}
\begin{tabulary}{\linewidth}{CCCCC}
\rowcolor{lightgray}
\toprule
               & Method & Unsup. Possible?  & UCF-Crime   & XD-Violence \\ \midrule

\multirow{9}{*}{\rotatebox[origin=c]{90}{Centralized}}        
               & Sultani \etal \cite{sultani2018real}    &  \ding{55} & 75.41    & - \\      
               & RTFM \cite{tian2021featuremagnitude}    &  \ding{55} & 84.30    & 89.34 \\   
               & MSL \cite{li2022self}    &  \ding{55} & 85.30    & - \\ 
               & S3R \cite{wu2022self}    &  \ding{55} & 85.99    & 53.52 \\
               & CLAWS+ \cite{zaheer2022clustering}    &  \ding{55} & 80.90    & - \\
               & PRV \cite{doshi2023privacy}    &  \ding{55} & \textbf{86.30}    & - \\
               & GCL \cite{zaheer2022generative}   &   \ding{51} &   79.84       &   82.18 \\
               & C2FPL \cite{anas2023c2fpl}    &  \ding{51}  & 83.40         & 89.34   \\
               & \our{}    &  \ding{51} & 85.50    & \textbf{90.04} \\
\midrule
\multirow{3}{*}{ \rotatebox[origin=c]{90}{Local}}
               & GCL \cite{zaheer2022generative}   &   \ding{51} &   65.32       &   59.91 \\
               & C2FPL \cite{anas2023c2fpl}    &  \ding{51}  & 65.85         & 63.4  \\
                & \our{}    &  \ding{51} & \textbf{67.47}    & \textbf{64.97}\\
\midrule
\multirow{4}{*}{\rotatebox[origin=c]{90}{Collab.}}              
               & PRV \cite{doshi2023privacy}    &  \ding{55} & 82.90    & - \\
               & GCL \cite{zaheer2022generative}   &   \ding{51} &   76.82       &   75.21 \\
               & C2FPL \cite{anas2023c2fpl}    &  \ding{51}  & 77.60         & 76.98   \\   
                & \our{}    &  \ding{51} & \textbf{83.23}    &  \textbf{85.67} \\
 \bottomrule
\end{tabulary}
\end{center}
\vspace{-20pt}
\caption{AUC performance comparisons of weakly supervision SOTA on UCF-Crime and XD-Violence datasets.}
\label{tab:ws_sota_comp}

\end{table}








\subsection{Collaborative Learning in VAD: A Case Study of Different Possible Scenarios}

In this section, we further explore the collaborative learning of video anomaly detection by proposing various scenarios of collaboration and consequent re-organization of the training data. Furthermore, we analyze and discuss the performance of \our{} under these scenarios.

\subsubsection{Training Data Splitting}

In real-world VAD applications, common sources of surveillance videos could be different government entities (e.g., department of transport or police) or private CCTV operating institutions (e.g., shopping malls or elderly-care facilities). \our{} is designed to enable collaborative learning of a joint anomaly detector between such data sources while eliminating the need to share training data. Considering this, we propose three different training data splits mimicking different kinds of collaborations between the participants. Each of these is explained below. 

\noindent\textbf{Random Split:}
A baseline setting where each participant has randomly distributed and equal number of anomalous \& normal videos for training. Figure \ref{fig:training_data_organization}a visualizes the random distribution of videos between ten participants on the UCF-crime dataset.

\noindent\textbf{Event Class Based Split:}
In this setting, each participant has training videos based on the anomalous events present within. For example, one participant may have road accident videos whereas another participant may have robbery videos. This setting is more challenging than random distribution because each participant may have a different number of videos. Figure \ref{fig:training_data_organization}b visualizes the distribution of videos between thirteen participants on the UCF-crime dataset.

\noindent\textbf{Scene Based Split:}
In this setting, each participant gets the videos based on the scenes/locations where the videos are recorded. For example, one participant may have surveillance videos for fuel stations, another participant may have indoor videos of offices, and so on. This is the most challenging and representative setting as the dataset is not balanced either among participants or within a participant for the normal and anomalous classes. Figure \ref{fig:training_data_organization}c visualizes the distribution of videos between eleven participants on the UCF-crime dataset.

The intuition behind these splits is that, in real-world scenarios, several participants for training a joint model may belong to different entities with different types of video data available at their disposal. For example, a police department may have videos related to street crimes, a children's daycare facility may have surveillance available for abuse or bullying, and a mall may have videos about stealing or shoplifting. More details on the data splitting strategies are provided in the Supplementary.


\subsubsection{Experiments Using Data Splits}
We conduct experiments on \our{} by splitting the UCF-crime dataset using the training data splits proposed (Figure \ref{fig:training_data_organization}) and report the results in Table \ref{tab:data_distribution_experiments}. For the random splitting, we vary the number of participants between $5$ and $50$ to additionally analyze the impact of the number of participants on training. Unsurprisingly, random split based training yields the highest AUC performance of 78.02\% with the participant number set to five. Experiments using event based splitting results in a closer performance of 77.35\%. This demonstrates that \our{} can efficiently handle data variations and partially imbalanced data among participants. With the most challenging training setting, scene based splitting, \our{} achieves an AUC of 73.99\%.

\renewcommand{\arraystretch}{1.1}
\begin{SCtable}
\footnotesize

\begin{tabulary}{\linewidth}{CCC}
\rowcolor{lightgray}
\toprule
\multicolumn{1}{c}{Split}     & Participants & AUC(\%) \\ \midrule
\multirow{4}{*}{Random (IID)} & 5                  & \textbf{78.02}    \\
                              & 10                 & 77.4    \\
                              & 25                 & 76.54   \\
                              & 50                 & 67.92   \\ \hline
\multicolumn{1}{c}{Event}     & 13                 & 77.35  \\\midrule
\multicolumn{1}{c}{Scene}     & 11                 & 73.99   \\ 
\bottomrule
\end{tabulary}
\caption{AUC \% performance of CLAP on UCF-Crime dataset using various proposed training splits under collaborative learning setting. }
\label{tab:data_distribution_experiments}
\end{SCtable}

\begin{figure}[t]
    \centering
    \includegraphics[width=1\linewidth]{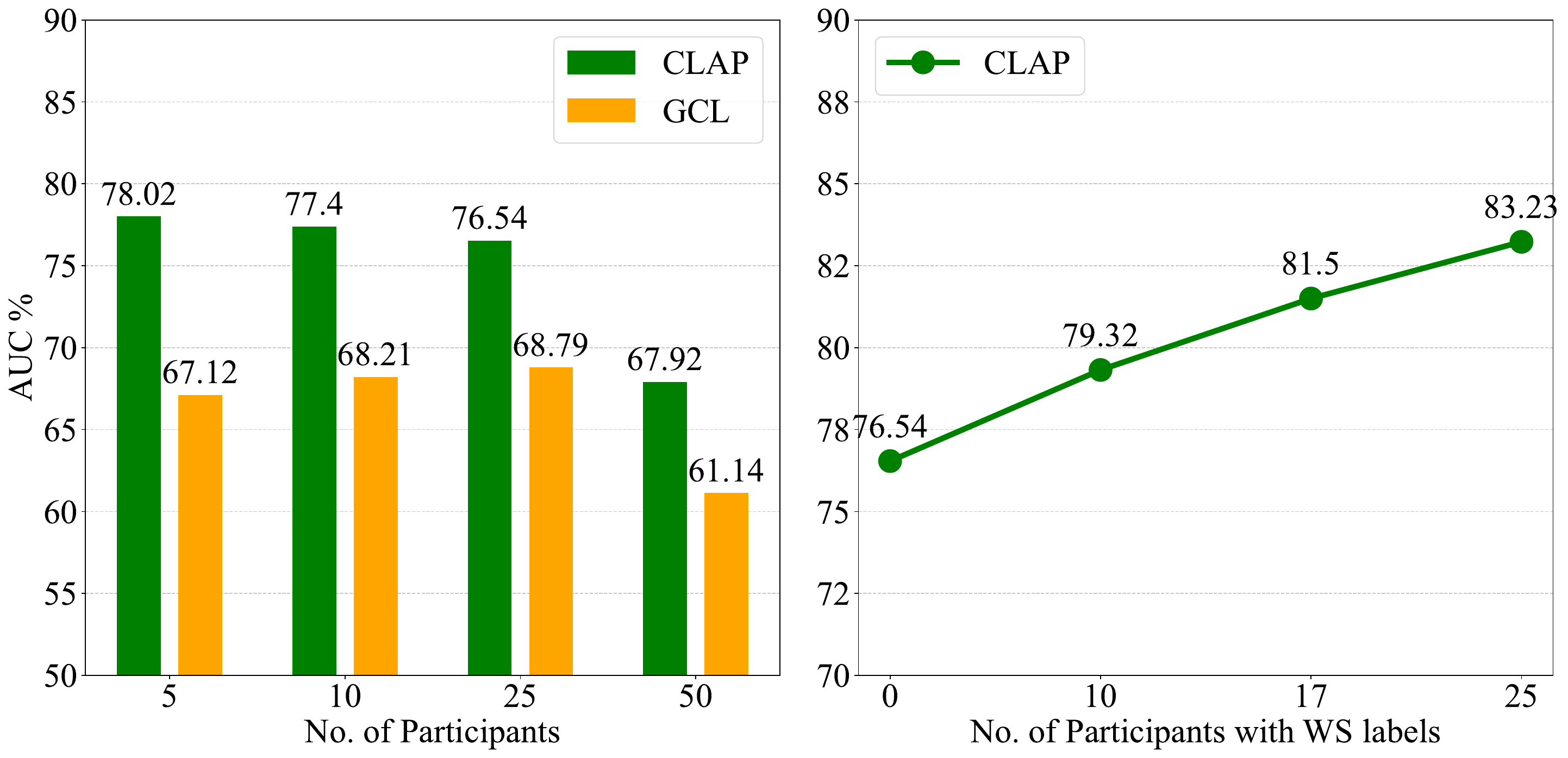}
    \caption{Left: Comparison between GCL \cite{zaheer2022generative} and CLAP with varying number of participants.  Right: Number of participants with weak video-level supervision available.}
    \label{fig:clapvs}
    \vspace{-5pt}
\end{figure}

\subsection{Analysis and Discussions}

\noindent\textbf{On Partial Weak-Supervision: }
The typical protocol for evaluating unsupervised methods under weakly-supervised settings is fairly simple. Once pseudo labels are generated for the whole dataset, for each video labeled as normal in the weakly supervised ground truth, label correction 
 on pseudo labels is applied before carrying out the training \cite{zaheer2022generative}. In the collaborative learning setting for real-world scenarios, ensuring the availability of any form of training labels means requiring each participant to annotate their data before participating in the collaborative training. While \our{} is fully unsupervised, meaning no labels are required for training, in this section we explore an interesting scenario where some of the participants may have labels available for training.
Figure \ref{fig:clapvs} (right) shows the results of \our{} on this setting using 25 participants. \our{} demonstrates consistent performance gains when more participants contribute with video-level labels towards the collaborative training.

\noindent\textbf{On Varying Number of Participants:}
To analyze the impact of varying numbers of participants on the unsupervised VAD training, we conduct a series of experiments using \our{} and GCL \cite{zaheer2022generative} with different numbers of participants and report the results in Figure \ref{fig:clapvs} (left). Overall, the performance stays comparable when the participant number is set to 5, 10, and 25. However, it drops notably when the participant number is set to 50. This may be attributed to the drop is the number of videos per participant, given that the dataset size remains the same. With a more large-scale dataset, our approach may be able to accommodate an even larger number of participants without dropping performance.


\begin{SCtable}
    
\centering
\scalebox{0.83}{
\begin{tabulary}{\linewidth}{CCCC}
\rowcolor{lightgray}
\toprule
FedAVG & SKA & PLR & AUC(\%) \\ \midrule
\ding{51}      & \ding{55}    &\ding{55}   & 76.2       \\ \midrule
\ding{51}     & \ding{51}    & \ding{55}   & 77.1       \\ \midrule
\ding{51}     & \ding{51}    & \ding{51}   & \textbf{78.02}       \\ \bottomrule

\end{tabulary}
}
\vspace{-7pt}
\caption{Ablation study of \our{} on UCF-Crime dataset. SKA: Server Knowledge Accumulation, PLR: Pseudo Label Refinement.}
\label{tab:ablation}

\end{SCtable}

\noindent\textbf{Ablation:}
To evaluate the contribution of the various components in \our, we conduct an ablation study and report the results in Table \ref{tab:ablation}. As seen, with each added component including SKA: Server Knowledge Accumulation stage and PLR: Psuedo-label refinement over the baseline training, notable performance gains are observed. This demonstrates the importance of all components proposed in \our{} towards unsupervised VAD.








\section{Conclusion}
\label{sec:conclusion}

We proposed a new baseline for anomaly detection capable of localizing anomalous events in a fully unsupervised fashion on a privacy-preserving collaborative learning configuration. We also introduced three new evaluation scenarios to extensively study anomaly detection approaches on various scenarios of collaborations and data availability. Using these scenarios, we evaluate our approach on two large-scale datasets including UCF-crime and XD-violence. A \textbf{limitation} of our approach is that the performance drops when the number of participants increases. Although some performance drop is expected in such a situation, we believe it is partly because of the limited training data available to each participant in this case. This can be addressed by curating large-scale anomaly detection datasets designed specifically for collaborative training.
{
    \small
    \bibliographystyle{ieeenat_fullname}
    \bibliography{main}

\begin{thebibliography}{42}
\providecommand{\natexlab}[1]{#1}
\providecommand{\url}[1]{\texttt{#1}}
\expandafter\ifx\csname urlstyle\endcsname\relax
  \providecommand{\doi}[1]{doi: #1}\else
  \providecommand{\doi}{doi: \begingroup \urlstyle{rm}\Url}\fi

\bibitem[Adnan et~al.(2022)Adnan, Kalra, Cresswell, Taylor, and Tizhoosh]{adnan2022federated}
Mohammed Adnan, Shivam Kalra, Jesse~C Cresswell, Graham~W Taylor, and Hamid~R Tizhoosh.
\newblock Federated learning and differential privacy for medical image analysis.
\newblock \emph{Scientific reports}, 12\penalty0 (1):\penalty0 1953, 2022.

\bibitem[Al-lahham et~al.(2023)Al-lahham, Tastan, Zaheer, and Nandakumar]{anas2023c2fpl}
Anas Al-lahham, Nurbek Tastan, Zaigham Zaheer, and Karthik Nandakumar.
\newblock A coarse-to-fine pseudo-labeling (c2fpl) framework for unsupervised video anomaly detection.
\newblock \emph{arXiv preprint arXiv:2310.17650}, 2023.

\bibitem[Almalik et~al.(2023)Almalik, Alkhunaizi, Almakky, and Nandakumar]{almalik2023fesvibs}
Faris Almalik, Naif Alkhunaizi, Ibrahim Almakky, and Karthik Nandakumar.
\newblock Fesvibs: Federated split learning of vision transformer with block sampling.
\newblock \emph{arXiv preprint arXiv:2306.14638}, 2023.

\bibitem[Bonawitz et~al.(2019)Bonawitz, Eichner, Grieskamp, Huba, Ingerman, Ivanov, Kiddon, Kone{\v{c}}n{\`y}, Mazzocchi, McMahan, et~al.]{bonawitz2019towards}
Keith Bonawitz, Hubert Eichner, Wolfgang Grieskamp, Dzmitry Huba, Alex Ingerman, Vladimir Ivanov, Chloe Kiddon, Jakub Kone{\v{c}}n{\`y}, Stefano Mazzocchi, Brendan McMahan, et~al.
\newblock Towards federated learning at scale: System design.
\newblock \emph{Proceedings of machine learning and systems}, 1:\penalty0 374--388, 2019.

\bibitem[Brik et~al.(2020)Brik, Ksentini, and Bouaziz]{brik2020federated}
Bouziane Brik, Adlen Ksentini, and Maha Bouaziz.
\newblock Federated learning for uavs-enabled wireless networks: Use cases, challenges, and open problems.
\newblock \emph{IEEE Access}, 8:\penalty0 53841--53849, 2020.

\bibitem[Darzidehkalani et~al.(2022)Darzidehkalani, Ghasemi-Rad, and van Ooijen]{darzidehkalani2022federated}
Erfan Darzidehkalani, Mohammad Ghasemi-Rad, and PMA van Ooijen.
\newblock Federated learning in medical imaging: part i: toward multicentral health care ecosystems.
\newblock \emph{Journal of the American College of Radiology}, 19\penalty0 (8):\penalty0 969--974, 2022.

\bibitem[Doshi and Yilmaz(2023)]{doshi2023privacy}
Keval Doshi and Yasin Yilmaz.
\newblock Privacy-preserving video understanding via transformer-based federated learning.
\newblock \emph{Data Science Conference}, 2023.

\bibitem[Du et~al.(2020)Du, Wu, Yoshinaga, Yau, Ji, and Li]{du2020federated}
Zhaoyang Du, Celimuge Wu, Tsutomu Yoshinaga, Kok-Lim~Alvin Yau, Yusheng Ji, and Jie Li.
\newblock Federated learning for vehicular internet of things: Recent advances and open issues.
\newblock \emph{IEEE Open Journal of the Computer Society}, 1:\penalty0 45--61, 2020.

\bibitem[Fantauzzo et~al.(2022)Fantauzzo, Fan{\`\i}, Caldarola, Tavera, Cermelli, Ciccone, and Caputo]{fantauzzo2022feddrive}
Lidia Fantauzzo, Eros Fan{\`\i}, Debora Caldarola, Antonio Tavera, Fabio Cermelli, Marco Ciccone, and Barbara Caputo.
\newblock Feddrive: Generalizing federated learning to semantic segmentation in autonomous driving.
\newblock In \emph{2022 IEEE/RSJ International Conference on Intelligent Robots and Systems (IROS)}, pages 11504--11511. IEEE, 2022.

\bibitem[Ghimire and Rawat(2022)]{ghimire2022recent}
Bimal Ghimire and Danda~B Rawat.
\newblock Recent advances on federated learning for cybersecurity and cybersecurity for federated learning for internet of things.
\newblock \emph{IEEE Internet of Things Journal}, 9\penalty0 (11):\penalty0 8229--8249, 2022.

\bibitem[Gong et~al.(2019)Gong, Liu, Le, Saha, Mansour, Venkatesh, and Hengel]{gong2019memorizing}
Dong Gong, Lingqiao Liu, Vuong Le, Budhaditya Saha, Moussa~Reda Mansour, Svetha Venkatesh, and Anton van~den Hengel.
\newblock Memorizing normality to detect anomaly: Memory-augmented deep autoencoder for unsupervised anomaly detection.
\newblock In \emph{Proceedings of the IEEE International Conference on Computer Vision}, pages 1705--1714, 2019.

\bibitem[Ionescu et~al.(2019)Ionescu, Khan, Georgescu, and Shao]{ionescu2019objectcentric}
Radu~Tudor Ionescu, Fahad~Shahbaz Khan, Mariana-Iuliana Georgescu, and Ling Shao.
\newblock Object-centric auto-encoders and dummy anomalies for abnormal event detection in video.
\newblock In \emph{Proceedings of the IEEE Conference on Computer Vision and Pattern Recognition}, pages 7842--7851, 2019.

\bibitem[Karimireddy et~al.(2020)Karimireddy, Kale, Mohri, Reddi, Stich, and Suresh]{karimireddy2020scaffold}
Sai~Praneeth Karimireddy, Satyen Kale, Mehryar Mohri, Sashank Reddi, Sebastian Stich, and Ananda~Theertha Suresh.
\newblock Scaffold: Stochastic controlled averaging for federated learning.
\newblock In \emph{International conference on machine learning}, pages 5132--5143. PMLR, 2020.

\bibitem[Lai et~al.(2022)Lai, Dai, Singapuram, Liu, Zhu, Madhyastha, and Chowdhury]{lai2022fedscale}
Fan Lai, Yinwei Dai, Sanjay Singapuram, Jiachen Liu, Xiangfeng Zhu, Harsha Madhyastha, and Mosharaf Chowdhury.
\newblock Fedscale: Benchmarking model and system performance of federated learning at scale.
\newblock In \emph{International Conference on Machine Learning}, pages 11814--11827. PMLR, 2022.

\bibitem[Li et~al.(2022)Li, Liu, and Jiao]{li2022self}
Shuo Li, Fang Liu, and Licheng Jiao.
\newblock Self-training multi-sequence learning with transformer for weakly supervised video anomaly detection.
\newblock In \emph{Proceedings of the AAAI Conference on Artificial Intelligence}, pages 1395--1403, 2022.

\bibitem[Li et~al.(2020)Li, Sahu, Zaheer, Sanjabi, Talwalkar, and Smith]{li2020federated}
Tian Li, Anit~Kumar Sahu, Manzil Zaheer, Maziar Sanjabi, Ameet Talwalkar, and Virginia Smith.
\newblock Federated optimization in heterogeneous networks.
\newblock \emph{Proceedings of Machine learning and systems}, 2:\penalty0 429--450, 2020.

\bibitem[Li et~al.(2021)Li, Tao, Zhang, Liu, and Xu]{li2021privacy}
Yijing Li, Xiaofeng Tao, Xuefei Zhang, Junjie Liu, and Jin Xu.
\newblock Privacy-preserved federated learning for autonomous driving.
\newblock \emph{IEEE Transactions on Intelligent Transportation Systems}, 23\penalty0 (7):\penalty0 8423--8434, 2021.

\bibitem[Liu et~al.(2020{\natexlab{a}})Liu, Huang, Luo, Huang, Liu, Chen, Feng, Chen, Yu, and Yang]{liu2020fedvision}
Yang Liu, Anbu Huang, Yun Luo, He Huang, Youzhi Liu, Yuanyuan Chen, Lican Feng, Tianjian Chen, Han Yu, and Qiang Yang.
\newblock Fedvision: An online visual object detection platform powered by federated learning.
\newblock In \emph{Proceedings of the AAAI conference on artificial intelligence}, pages 13172--13179, 2020{\natexlab{a}}.

\bibitem[Liu et~al.(2020{\natexlab{b}})Liu, Peng, Kang, Iliyasu, Niyato, and Abd El-Latif]{liu2020secure}
Yi Liu, Jialiang Peng, Jiawen Kang, Abdullah~M Iliyasu, Dusit Niyato, and Ahmed~A Abd El-Latif.
\newblock A secure federated learning framework for 5g networks.
\newblock \emph{IEEE Wireless Communications}, 27\penalty0 (4):\penalty0 24--31, 2020{\natexlab{b}}.

\bibitem[McMahan et~al.(2017)McMahan, Moore, Ramage, Hampson, and y~Arcas]{mcmahan2017communication}
Brendan McMahan, Eider Moore, Daniel Ramage, Seth Hampson, and Blaise~Aguera y Arcas.
\newblock Communication-efficient learning of deep networks from decentralized data.
\newblock In \emph{Artificial intelligence and statistics}, pages 1273--1282. PMLR, 2017.

\bibitem[Mothukuri et~al.(2021)Mothukuri, Parizi, Pouriyeh, Huang, Dehghantanha, and Srivastava]{mothukuri2021survey}
Viraaji Mothukuri, Reza~M Parizi, Seyedamin Pouriyeh, Yan Huang, Ali Dehghantanha, and Gautam Srivastava.
\newblock A survey on security and privacy of federated learning.
\newblock \emph{Future Generation Computer Systems}, 115:\penalty0 619--640, 2021.

\bibitem[Ng et~al.(2021)Ng, Lan, Yao, Chan, and Feng]{ng2021federated}
Dianwen Ng, Xiang Lan, Melissa Min-Szu Yao, Wing~P Chan, and Mengling Feng.
\newblock Federated learning: a collaborative effort to achieve better medical imaging models for individual sites that have small labelled datasets.
\newblock \emph{Quantitative Imaging in Medicine and Surgery}, 11\penalty0 (2):\penalty0 852, 2021.

\bibitem[Nguyen et~al.(2022)Nguyen, Do, Tran, Nguyen, Duong, Phan, Tjiputra, and Tran]{nguyen2022deep}
Anh Nguyen, Tuong Do, Minh Tran, Binh~X Nguyen, Chien Duong, Tu Phan, Erman Tjiputra, and Quang~D Tran.
\newblock Deep federated learning for autonomous driving.
\newblock In \emph{2022 IEEE Intelligent Vehicles Symposium (IV)}, pages 1824--1830. IEEE, 2022.

\bibitem[Pang et~al.(2023)Pang, Ni, and Zhong]{pang2023federated}
Yiran Pang, Zhen Ni, and Xiangnan Zhong.
\newblock Federated learning for crowd counting in smart surveillance systems.
\newblock \emph{IEEE Internet of Things Journal}, 2023.

\bibitem[Petz(2001)]{petz2001entropy}
D{\'e}nes Petz.
\newblock Entropy, von neumann and the von neumann entropy: Dedicated to the memory of alfred wehrl.
\newblock In \emph{John von Neumann and the foundations of quantum physics}, pages 83--96. Springer, 2001.

\bibitem[Purwanto et~al.(2021)Purwanto, Chen, and Fang]{Purwanto2021ICCV}
Didik Purwanto, Yie-Tarng Chen, and Wen-Hsien Fang.
\newblock Dance with self-attention: A new look of conditional random fields on anomaly detection in videos.
\newblock In \emph{Proceedings of the IEEE/CVF International Conference on Computer Vision (ICCV)}, pages 173--183, 2021.

\bibitem[Sabokrou et~al.(2018)Sabokrou, Khalooei, Fathy, and Adeli]{sabokrou2018ALOCC}
Mohammad Sabokrou, Mohammad Khalooei, Mahmood Fathy, and Ehsan Adeli.
\newblock Adversarially learned one-class classifier for novelty detection.
\newblock In \emph{Proceedings of the IEEE Conference on Computer Vision and Pattern Recognition}, pages 3379--3388, 2018.

\bibitem[Sada et~al.(2019)Sada, Bouras, Ma, Runhe, and Ning]{sada2019distributed}
Abdelkarim~Ben Sada, Mohammed~Amine Bouras, Jianhua Ma, Huang Runhe, and Huansheng Ning.
\newblock A distributed video analytics architecture based on edge-computing and federated learning.
\newblock In \emph{2019 IEEE Intl Conf on Dependable, Autonomic and Secure Computing, Intl Conf on Pervasive Intelligence and Computing, Intl Conf on Cloud and Big Data Computing, Intl Conf on Cyber Science and Technology Congress (DASC/PiCom/CBDCom/CyberSciTech)}, pages 215--220. IEEE, 2019.

\bibitem[Sultani et~al.(2018)Sultani, Chen, and Shah]{sultani2018real}
Waqas Sultani, Chen Chen, and Mubarak Shah.
\newblock Real-world anomaly detection in surveillance videos.
\newblock In \emph{Proceedings of the IEEE conference on computer vision and pattern recognition}, pages 6479--6488, 2018.

\bibitem[Tang et~al.(2022)Tang, Hu, and Xu]{tang2022federated}
Zhongyun Tang, Haiyang Hu, and Chonghuan Xu.
\newblock A federated learning method for network intrusion detection.
\newblock \emph{Concurrency and Computation: Practice and Experience}, 34\penalty0 (10):\penalty0 e6812, 2022.

\bibitem[Tian et~al.(2021{\natexlab{a}})Tian, Pang, Chen, Singh, Verjans, and Carneiro]{tian2021featuremagnitude}
Yu Tian, Guansong Pang, Yuanhong Chen, Rajvinder Singh, Johan~W Verjans, and Gustavo Carneiro.
\newblock Weakly-supervised video anomaly detection with robust temporal feature magnitude learning.
\newblock \emph{arXiv preprint arXiv:2101.10030}, 2021{\natexlab{a}}.

\bibitem[Tian et~al.(2021{\natexlab{b}})Tian, Pang, Chen, Singh, Verjans, and Carneiro]{tian2021weakly}
Yu Tian, Guansong Pang, Yuanhong Chen, Rajvinder Singh, Johan~W Verjans, and Gustavo Carneiro.
\newblock Weakly-supervised video anomaly detection with robust temporal feature magnitude learning.
\newblock In \emph{Proceedings of the IEEE/CVF international conference on computer vision}, pages 4975--4986, 2021{\natexlab{b}}.

\bibitem[Wu et~al.(2022)Wu, Hsieh, Chen, Fuh, and Liu]{wu2022self}
Jhih-Ciang Wu, He-Yen Hsieh, Ding-Jie Chen, Chiou-Shann Fuh, and Tyng-Luh Liu.
\newblock Self-supervised sparse representation for video anomaly detection.
\newblock In \emph{Computer Vision--ECCV 2022: 17th European Conference, Tel Aviv, Israel, October 23--27, 2022, Proceedings, Part XIII}, pages 729--745. Springer, 2022.

\bibitem[Wu et~al.(2020)Wu, Liu, Shi, Sun, Shao, Wu, and Yang]{wu2020not}
Peng Wu, Jing Liu, Yujia Shi, Yujia Sun, Fangtao Shao, Zhaoyang Wu, and Zhiwei Yang.
\newblock Not only look, but also listen: Learning multimodal violence detection under weak supervision.
\newblock In \emph{Computer Vision--ECCV 2020: 16th European Conference, Glasgow, UK, August 23--28, 2020, Proceedings, Part XXX 16}, pages 322--339. Springer, 2020.

\bibitem[Zaheer et~al.(2020{\natexlab{a}})Zaheer, Lee, Astrid, and Lee]{zaheer2020old}
Muhammad~Zaigham Zaheer, Jin-ha Lee, Marcella Astrid, and Seung-Ik Lee.
\newblock Old is gold: Redefining the adversarially learned one-class classifier training paradigm.
\newblock In \emph{Proceedings of the IEEE/CVF Conference on Computer Vision and Pattern Recognition}, pages 14183--14193, 2020{\natexlab{a}}.

\bibitem[Zaheer et~al.(2020{\natexlab{b}})Zaheer, Mahmood, Astrid, and Lee]{zaheer2020claws}
Muhammad~Zaigham Zaheer, Arif Mahmood, Marcella Astrid, and Seung-Ik Lee.
\newblock Claws: Clustering assisted weakly supervised learning with normalcy suppression for anomalous event detection.
\newblock In \emph{Computer Vision--ECCV 2020: 16th European Conference, Glasgow, UK, August 23--28, 2020, Proceedings, Part XXII 16}, pages 358--376. Springer, 2020{\natexlab{b}}.

\bibitem[Zaheer et~al.(2020{\natexlab{c}})Zaheer, Mahmood, Shin, and Lee]{zaheer2020self}
Muhammad~Zaigham Zaheer, Arif Mahmood, Hochul Shin, and Seung-Ik Lee.
\newblock A self-reasoning framework for anomaly detection using video-level labels.
\newblock \emph{IEEE Signal Processing Letters}, 27:\penalty0 1705--1709, 2020{\natexlab{c}}.

\bibitem[Zaheer et~al.(2022{\natexlab{a}})Zaheer, Mahmood, Astrid, and Lee]{zaheer2022clustering}
Muhammad~Zaigham Zaheer, Arif Mahmood, Marcella Astrid, and Seung-Ik Lee.
\newblock Clustering aided weakly supervised training to detect anomalous events in surveillance videos, 2022{\natexlab{a}}.

\bibitem[Zaheer et~al.(2022{\natexlab{b}})Zaheer, Mahmood, Khan, Segu, Yu, and Lee]{zaheer2022generative}
M~Zaigham Zaheer, Arif Mahmood, M~Haris Khan, Mattia Segu, Fisher Yu, and Seung-Ik Lee.
\newblock Generative cooperative learning for unsupervised video anomaly detection.
\newblock In \emph{Proceedings of the IEEE/CVF Conference on Computer Vision and Pattern Recognition}, pages 14744--14754, 2022{\natexlab{b}}.

\bibitem[Zhang et~al.(2023{\natexlab{a}})Zhang, Li, Qi, Wang, Qing, Huang, and Yang]{zhang2023exploiting}
Chen Zhang, Guorong Li, Yuankai Qi, Shuhui Wang, Laiyun Qing, Qingming Huang, and Ming-Hsuan Yang.
\newblock Exploiting completeness and uncertainty of pseudo labels for weakly supervised video anomaly detection.
\newblock In \emph{Proceedings of the IEEE/CVF Conference on Computer Vision and Pattern Recognition}, pages 16271--16280, 2023{\natexlab{a}}.

\bibitem[Zhang et~al.(2021)Zhang, Bosch, and Olsson]{zhang2021end}
Hongyi Zhang, Jan Bosch, and Helena~Holmstr{\"o}m Olsson.
\newblock End-to-end federated learning for autonomous driving vehicles.
\newblock In \emph{2021 International Joint Conference on Neural Networks (IJCNN)}, pages 1--8. IEEE, 2021.

\bibitem[Zhang et~al.(2023{\natexlab{b}})Zhang, Vahidian, Kuo, Li, Zhang, Wang, and Chen]{zhang2023towards}
Jianyi Zhang, Saeed Vahidian, Martin Kuo, Chunyuan Li, Ruiyi Zhang, Guoyin Wang, and Yiran Chen.
\newblock Towards building the federated gpt: Federated instruction tuning.
\newblock \emph{arXiv preprint arXiv:2305.05644}, 2023{\natexlab{b}}.

\end{thebibliography}
}

\clearpage
\setcounter{page}{1}
\maketitlesupplementary


\begin{figure}[t]
    \centering
    \includegraphics[width=1\linewidth]{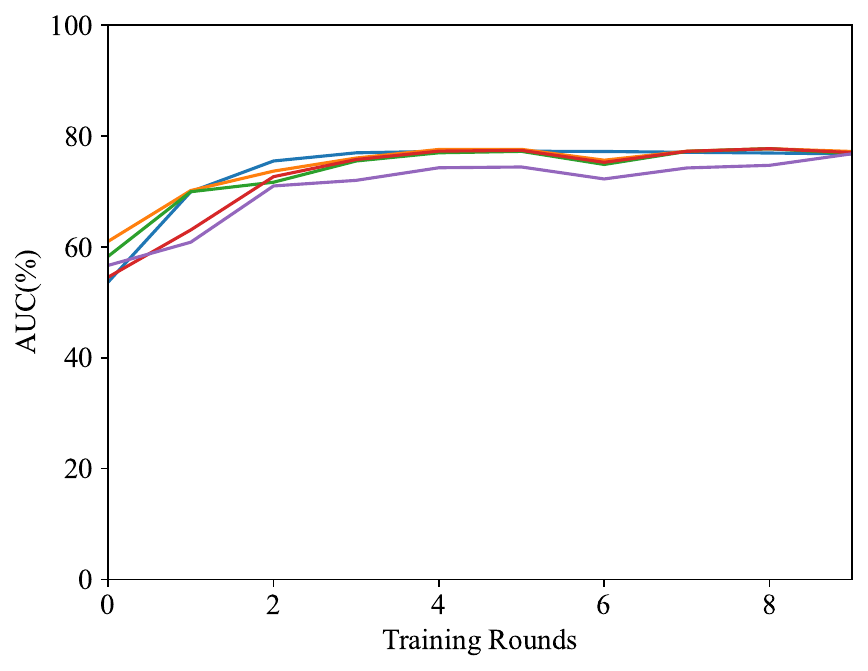}
    \caption{Empirical training convergence. Experiments are run using 5 different seeds enabling randomized data splits across participants. \our{} achieves an average AUC of $77.32 \pm 0.189$. }
    \label{fig:conv}
\end{figure}

\begin{figure*}[t]
    \centering    \includegraphics[width=0.75\linewidth]{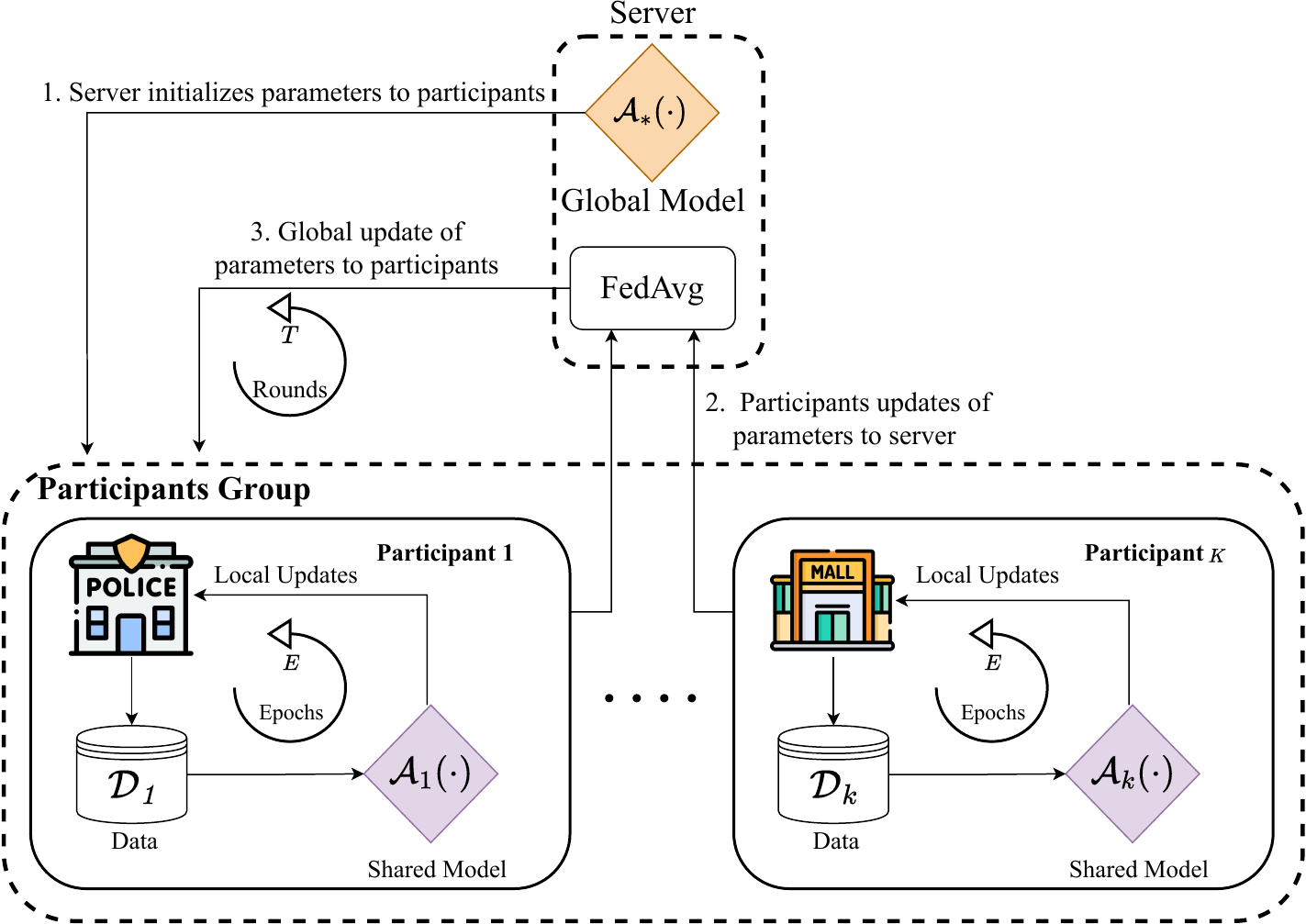}
    \caption{Abstract level Flowchart of our collaborative training scheme.}
    \label{fig:enter-label}
\end{figure*}

\section{Collabroative algorithms Study}
\label{otherfl}
In addition to the main results in the manuscript where FedAVG is used as the main FL method, we conduct experiments using other FL methods including FedProx \cite{li2020federated} and SCAFFOLD \cite{karimireddy2020scaffold}. In these experiments, CLAP achievs 73.4\% \& 73.7\% AUC respectively on \textbf{scene based split}. Overall, the performance is comparable with the 73.99\% AUC when using FedAVG. 

\section{Training convergence}

As \our is an unsupervised learning model where each participant uses its share of the data to collaborate towards training a joint model, we empirically validate its convergence by repeating the training on UCF-Crime using 5 different random seeds. These random seeds also enable random data splits between the participants. As seen in Figure \ref{fig:conv}, \our achieves an average AUC of 77.32\% $\pm$ 0.189\%. This shows the robustness of \our in yielding good performance with small variation even with significant variation in the dataset splits across participants.


\section{Bandwidth Consumption}
In real-world surveillance applications, network bandwidth allowing data communication between the training server and the participants can be limited due to several factors such as remote locations, large number of participants, etc. Given the involvement of lengthy surveillance videos for anomaly detection, a collaborative learning approach such as \our should preferably communicate a limited amount of data per training round.
As shown in Algorithm \ref{alg:flad}, the server receives the Gaussian parameters from each participant in addition to receiving the gradients of each local model (2.1 M parameters) during the training rounds. Therefore, on each communication round, \our communicates an average of 6.07 Mega Bytes (MB) from each participant. Given 10 training rounds, the overall data transfer remains around 60.7 MB which is significantly lower than the case of central training where all data is transferred to the central server for training. 


%


\section{Dataset Splitting Strategies}
As described in Section 4.5 of the manuscript, collaborative learning in video anomaly detection (VAD) may have several possible scenarios. Careful consideration of these scenarios leads to three different data splitting strategies including random, event class, and scene-based.
Each of these is explained further next:

\noindent\textbf{Random Split:}
Random Split is a baseline strategy where each participant is assigned videos randomly while ensuring a comparable number of normal and anomalous videos.
Example visualizations of some videos taken from a single participant are provided in Figure \ref{fig:random_split}.


\noindent\textbf{Event Class Based Split:}
Each anomalous activity can be classified into different categories of events, e.g., road accidents, robbery, fighting, shooting, or riots. The intuition behind this split is that each collaborating participant may have a certain type of anomalous examples. A better performance of an anomaly detection network on this setting may indicate the success of collaborative learning between different organizations contributing videos containing different types of anomalous events from each other. This setting is more challenging than random distribution because each participant may have a different number of videos containing certain events. Example visualizations of some videos taken from a single participant are provided in Figure \ref{fig:event_split}.


\noindent\textbf{Scene Based Split:}
In this setting, each participant is assumed to have videos based on the scenes/locations where the videos are recorded. For example, one participant may have surveillance videos for fuel stations, another participant may have indoor videos of offices, and so on.
The intuition behind this split is that 
similar anomalous events may occur at different locations and captured by different participants. 
This is the most challenging setting as the dataset is not balanced either among participants or within a participant for the normal and anomalous classes.
Example visualizations of some videos taken from a single participant are provided in Figure \ref{fig:scene_split}.

\section{Architecture and Implementation Details}

\begin{figure*}[t]
\begin{center}
    \includegraphics[width=0.8\linewidth]{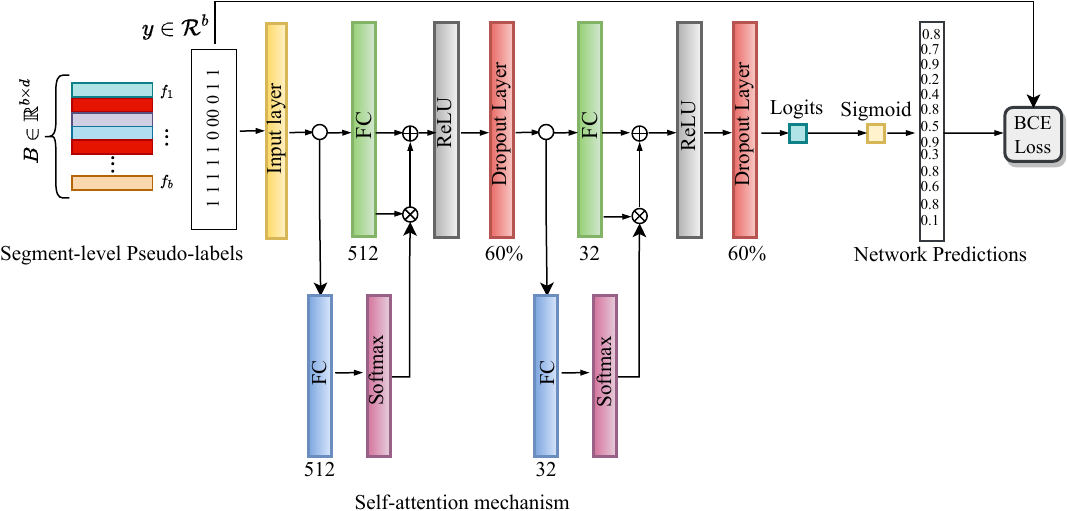}
\end{center}
   \caption{Learning network used in \our: The training batch containing pseudo-labeled feature vectors is the input to the FC backbone network (upper). In addition to the backbone network, we add two self-attention layers (lower).} 
\label{fig:architecture}
\end{figure*}

Our learning network, as seen in Figure \ref{fig:architecture},
consists of a fully connected (FC) network and two self-attention layers. The FC network has two fully connected layers and one output layer for binary classification. A ReLU activation function and a dropout layer follow each FC layer. The FC layers have 512 and 32 neurons respectively. The self-attention layers, with dimensions matching their respective FC layers, are followed by a Softmax activation function. Unlike previous works \cite{zaheer2020claws,zaheer2022clustering}, we compute Softmax probabilities over the feature dimension instead of the batch size dimension. The final anomaly score prediction ranging $[0,1]$ in our network is obtained through a Sigmoid activation function in the output layer. We use binary cross-entropy loss along with $L_{2}$ regularizer as our training loss function.

\section{Datasets}

Two large-scale video anomaly detection benchmark datasets are used to evaluate our approach: UCF-Crime \cite{sultani2018real} and XD-Violence \cite{wu2020not}. These datasets are originally labeled for weakly supervised VAD tasks, where video-level labels are present for training and frame-level labels are provided only for testing. In our unsupervised VAD experiments, we completely discard the provided labels before carrying out the training.

\subsection{UCF-Crime}
\begin{figure*}[t]
\begin{center}
    \includegraphics[width=1\linewidth]{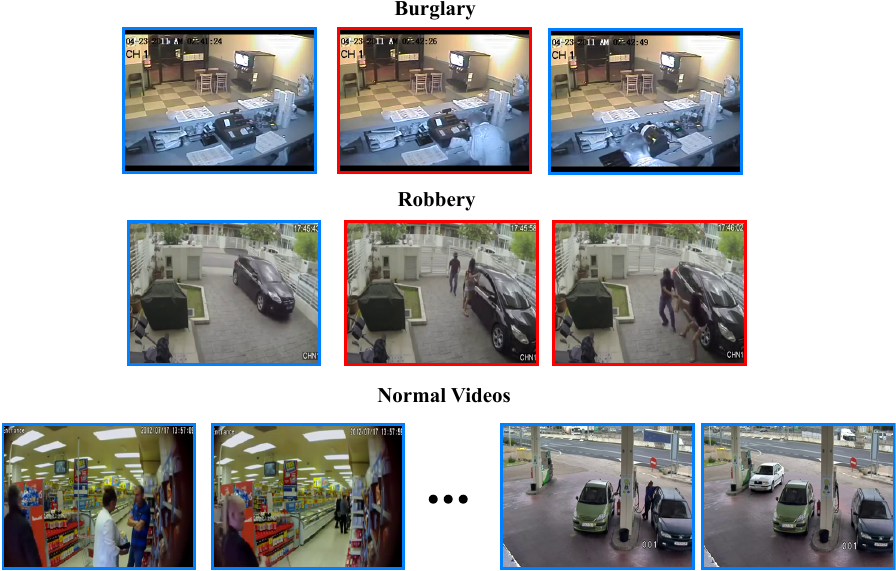}
\end{center}
   \caption{Example of UCF-crime videos in the random split taken from one of the participants. The blue borders represent normal events while the red borders represent anomalous events.} 
\label{fig:random_split}
\end{figure*}

\begin{figure*}[t]
\begin{center}
    \includegraphics[width=1\linewidth]{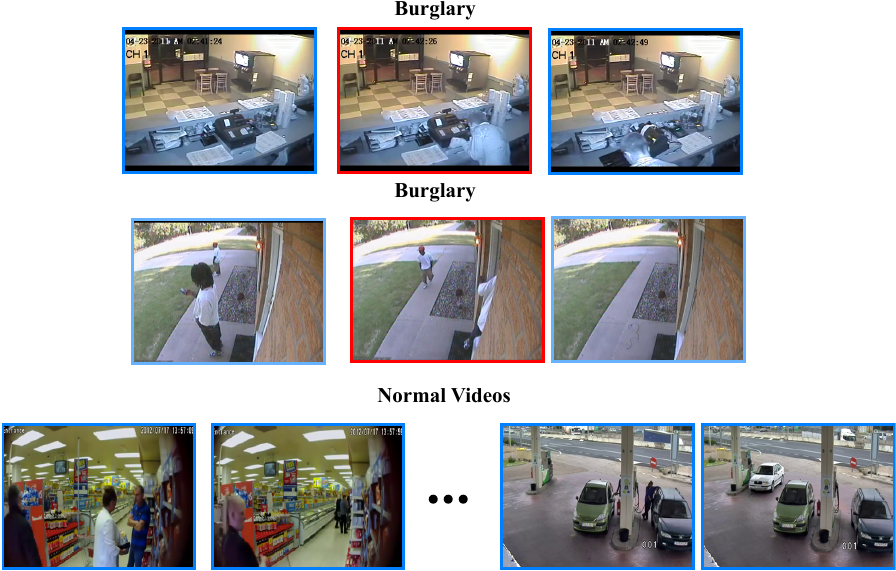}
\end{center}
   \caption{Example of UCF-crime videos in the event-based split taken from one of the participants. For each participant, anomalous events are the same but the background scenes can be different. The blue borders represent normal events while the red borders represent anomalous events.} 
\label{fig:event_split}
\end{figure*}

\begin{figure*}[t]
\begin{center}
    \includegraphics[width=0.7\linewidth]{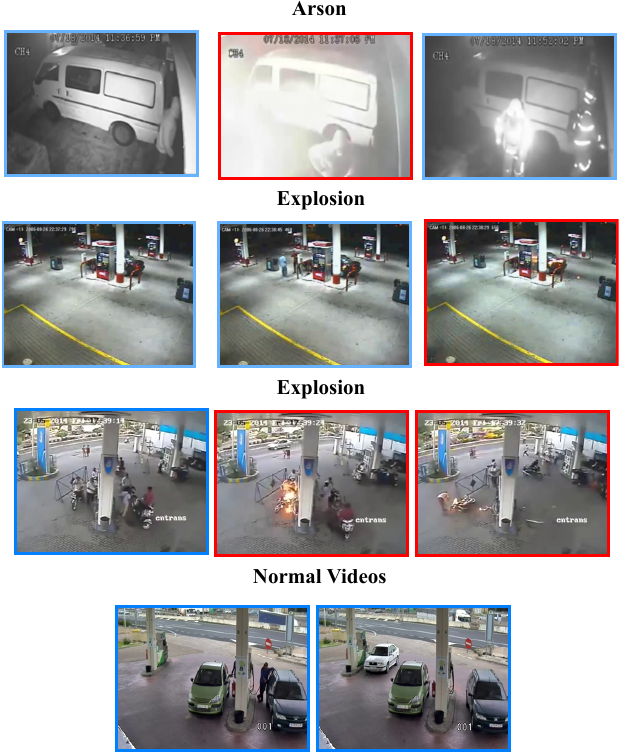}
\end{center}
   \caption{Example of UCF-crime videos in the scene-based split taken from a participant having videos of fuel pumps and automotive workshops). For each participant, anomalous events can be different but the overall background scenes are similar. The blue borders represent normal events while the red borders represent anomalous events.} 
\label{fig:scene_split}
\end{figure*}

UCF-Crime consists of 1,610 training videos and 290 testing videos covering 13 anomaly categories including Abuse, Arrest, Arson, Assault, ... etc. Some examples of these videos are shown in Figures \ref{fig:random_split}, \ref{fig:event_split}, \& \ref{fig:scene_split}. These videos were gathered from actual surveillance camera feeds, amounting to a combined duration of 128 hours.

\subsection{XD-Violence}
XD-Violence is a multi-modal dataset sourced from various channels, including sports streaming videos, movies, and web videos. The dataset encompasses a total of 3,954 training videos and 800 testing videos. These videos collectively span approximately 217 hours.

\balance

\end{document}